\documentclass[10pt,twocolumn,letterpaper]{article}
\usepackage{multirow}
\usepackage{iccv}
\usepackage{times}
\usepackage{epsfig}

\usepackage{amsmath}
\usepackage{amssymb}
\usepackage{graphicx}

\usepackage{booktabs}
\usepackage[accsupp]{axessibility}

\usepackage{caption}
\usepackage{subcaption}
\usepackage{multirow}
\usepackage{iccv}
\usepackage{times}
\usepackage{epsfig}
\usepackage{booktabs}
\usepackage[T1]{fontenc}
\usepackage[utf8]{inputenc}
\usepackage{authblk}

\usepackage[colorlinks,
            linkcolor=red,
            anchorcolor=blue,
            citecolor=green
            ]{hyperref}

\iccvfinalcopy 

\def\iccvPaperID{****} 

\ificcvfinal\fi

\begin{document}

\title{MB-TaylorFormer: Multi-branch Efficient Transformer Expanded by Taylor Formula for Image Dehazing}


\author{
Yuwei Qiu$^{1}$ \ \  Kaihao Zhang$^{2}$ \ \ Chenxi Wang$^{1}$ \ \ Wenhan Luo$^1$ \ \  Hongdong Li$^2$ \ \  Zhi Jin$^{1,3}$\thanks{Corresponding author: jinzh26@mail2.sysu.edu.cn}  \\
$^1$ Sun Yat-sen University \ \ \ \ \ \ \ \ \ \ 
$^2$ Australian National University \\
$^3$ Guangdong Provincial Key Laboratory of Robotics and Digital Intelligent Manufacturing Technology
}


\maketitle
\ificcvfinal\thispagestyle{empty}\fi

\begin{abstract}
 In recent years, Transformer networks are beginning to replace pure convolutional neural networks (CNNs) in the field of computer vision due to their global receptive field and adaptability to input. However, the quadratic computational complexity of softmax-attention limits the wide application in image dehazing task, especially for high-resolution images. To address this issue, we propose a new Transformer variant, which applies the Taylor expansion to approximate the softmax-attention and achieves linear computational complexity. A multi-scale attention refinement module is proposed as a complement to correct the error of the Taylor expansion. Furthermore, we introduce a multi-branch architecture with multi-scale patch embedding to the proposed Transformer, which embeds features by overlapping deformable convolution of different scales. The design of multi-scale patch embedding is based on three key ideas: 1) various sizes of the receptive field; 2) multi-level semantic information; 3) flexible shapes of the receptive field. Our model, named Multi-branch Transformer expanded by Taylor formula (MB-TaylorFormer), can embed coarse to fine features more flexibly at the patch embedding stage and capture long-distance pixel interactions with limited computational cost. Experimental results on several dehazing benchmarks show that MB-TaylorFormer achieves state-of-the-art (SOTA) performance with a light computational burden. The source code and pre-trained models are available at \href{https://github.com/FVL2020/ICCV-2023-MB-TaylorFormer}{\textcolor{blue}{https://github.com/FVL2020/ICCV-2023-MB-TaylorFormer}}.
\end{abstract}

\begin{figure}[t]
\centering
    \includegraphics[width=0.42\textwidth]{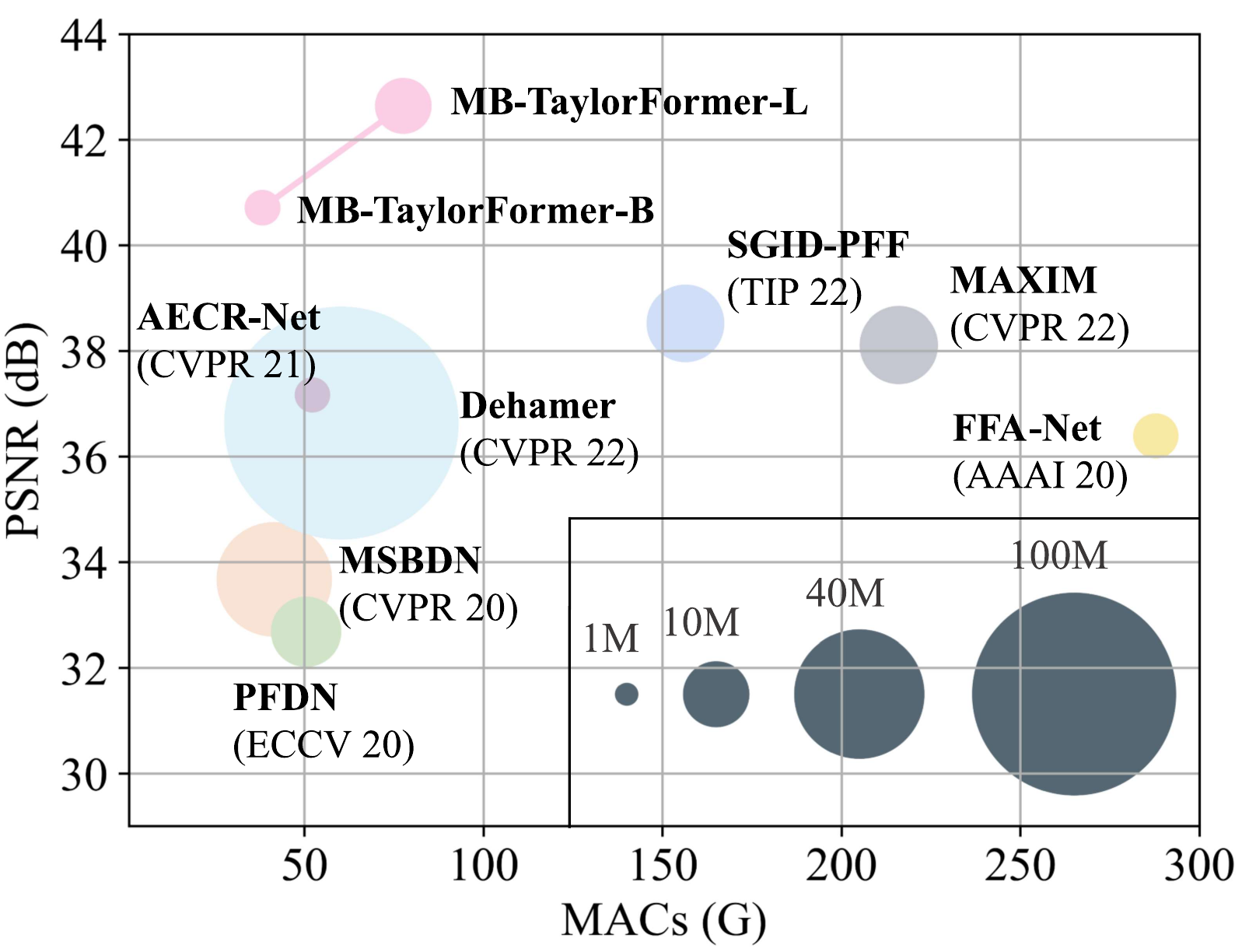}
\vspace{-0.2cm}

  \caption{\textbf{Improvement of MB-TaylorFormer over the SOTA approaches.} The circle size is proportional to the number of model parameters. All models are trained on SOTS-Indoor~\cite{8451944}.}
    \label{fig:figure1}
\vspace{-0.3cm}
\end{figure}
\section{Introduction}

Single image dehazing is an image restoration task which aims to estimate latent haze-free images from hazy images. Starting with early CNN-based approaches~\cite{ren2016single,cai2016dehazenet} and their revolutionary performance in dehazing, haze removal gradually shifts from prior based strategies~\cite{zhu2015fast,he2010single} to deep learning-based methods. In the past decade, deep dehazing networks achieve significant performance improvement due to advanced architectures like multi-scale information fusion~\cite{ren2016single,qu2019enhanced,liu2019griddehazenet}, sophisticated variants of convolution~\cite{wu2021contrastive,liu2020trident}, and attention mechanisms~\cite{qin2020ffa,zhang2020pyramid,hong2020distilling}.

 Recently, Transformer has been popularly employed in various computer vision tasks and subsequently contributes greatly to the progress of high-level vision tasks~\cite{liu2021swin,wang2021pyramid,touvron2021training,carion2020end}. In the field of image dehazing, there are several challenges with the direct application of Transformer: 1) the computational complexity of Transformer is quadratic with the resolution of feature map, which makes it poorly suited to the pixel-to-pixel task of dehazing. Although some works apply self-attention in small spatial windows~\cite{liang2021swinir,wang2022uformer} to relieve this problem, the receptive field of Transformer is restricted; 2) the basic elements of visual Transformer usually have more flexible scales~\cite{liu2021swin}. However, existing visual Transformer networks~\cite{zamir2022restormer,wang2022uformer} generally generate fixed-scale tokens by fixed convolution kernels. Thus, there is still room for improvement via introducing flexible patch embedding Transformer to the dehazing task.

 To address the first challenge, we propose a Transformer variant expanded by Taylor formula (TaylorFormer), which applies self-attention on the entire feature map across spatial dimension and maintains linear computational complexity. Specifically, we calculate the weights of self-attention by performing a Taylor expansion on softmax, and then reduce the computational complexity of self-attention from $\mathcal{O}(n^{2})$ to $\mathcal{O}(n)$ by applying the associative law of matrix multiplication. This strategy brings three advantages: 1) it retains the ability of Transformer to model long-distance dependencies among data, and avoids reducing the receptive field caused by window splitting~\cite{song2022vision}; 2) it provides a stronger value approximation than methods using a kernel-based formulation of self-attention~\cite{katharopoulos2020transformers}, and is similar to the vanilla Transformer~\cite{vaswani2017attention}; 3) it makes the Transformer concerned with pixel-level interactions rather than channel-level interactions~\cite{zamir2022restormer}, which allows for more fine-grained processing of features.

  Considering the error caused by ignoring Peano's form of remainder~\cite{peano1889nouvelle} in the Taylor formula, we introduce a multi-scale attention refinement (MSAR) module to refine TaylorFormer. We exploit the local correlation within image by convolving the local information of queries and keys to output a feature map with scaling factors. The number of feature map channels  is equal to the number of heads in multi-head self-attention (MSA), so each head has a corresponding scaling factor. Our experiments show that the proposed MSAR module effectively improves model performance with tiny computational burden (see Table~\ref{tab:tab4}).
    
To tackle the second challenge, inspired by the success of inception modules~\cite{szegedy2016rethinking,szegedy2015going} and deformable convolutions~\cite{dai2017deformable} in CNN-based dehazing networks~\cite{goncalves2017deepdive,wang2018aipnet,perez2021ntire,wu2021contrastive}, we propose a multi-branch encoder-decoder backbone for TaylorFormer, termed ad MB-TaylorForemr, based on multi-scale patch embedding. The multi-scale patch embedding has various sizes of receptive field, multi-level semantic information, and flexible shape of receptive field. Considering that the generation of each token should follow the local relevance prior, we truncate the offsets of the deformable convolutions. We reduce computational complexity and the number of parameters by the depthwise separable method. The tokens from different scales are then fed independently into TaylorFormer and finally fused. The multi-scale patch embedding module is capable of generating tokens with different scales and dimensions, and the multi-branch structure is capable of processing them simultaneously to capture more powerful features.

To summarize, our main contributions are as follows: (1) We propose a new variant of linearized Transformer based on Taylor expansion to model long-distance interactions between pixels without window splitting. An MSAR module is introduced to further correct errors in the self-attention of TaylorFormer; (2) We design a multi-branch architecture with multi-scale patch embedding. Among it, multiple field sizes, flexible shape of receptive field, and multi-level semantic information can help simultaneously generate tokens with multi-scales and capture more powerful features; (3) Experimental results on public synthetic and real dehazing datasets show that the proposed MB-TaylorForme achieves state-of-the-art (SOTA) performance with few parameters and MACs.
    

\section{Related Works}
\begin{figure*}[htbp]
    \centering
    \includegraphics[width=17cm]{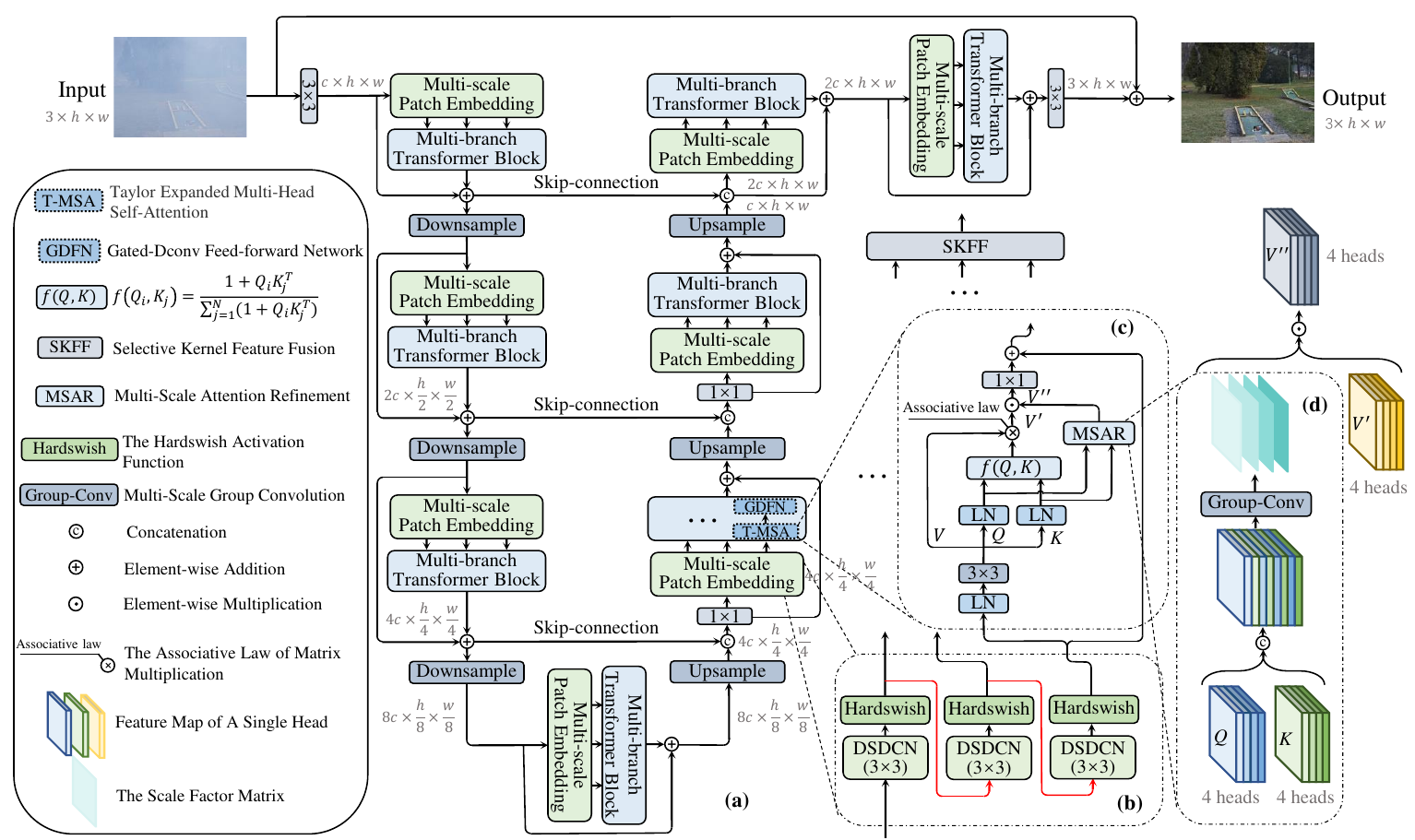}
    \caption{\textbf{Architecture of MB-TaylorFormer.} (a) MB-TaylorFormer consists of a multi-branch hierarchical design based on multi-scale patch embedding. (b) Multi-scale patch embedding embeds coarse-to-fine patches. (c) TaylorFormer with linear computational complexity. (d) MSAR module compensates for errors in Taylor expansion.} 
    \label{fig:pipeline}
\vspace{-0.1cm}
\end{figure*} 
    \noindent\textbf{CNN-based Image Dehazing}. With the development of CNN, significant progress has been achieved in image restoration tasks \cite{wang2023gridformer,zhang2022deep,zhang2022enhanced,zhang2020deblurring,zhang2021deep,wang2023fourllie,wang2023brightenandcolorize}, including image dehazing \cite{cai2016dehazenet,qin2020ffa,guo2022image,bai2022self,zheng2022t,wang2022restoring}.
    %
    One way is based on the atmospheric scattering model, such as DehazeNet~\cite{cai2016dehazenet}, DCPDN~\cite{zhang2018densely}, and Aod-Net~\cite{li2017aod}. However, the performance is hardly satisfactory when the atmospheric scattering model fails. The other way is based on the idea of image conversion, which is not dependent on the atmospheric scattering model. These kinds of methods predict haze-free images in an end-to-end manner, such as EPDN~\cite{qu2019enhanced}, FFA-Net~\cite{qin2020ffa} and AECRNet~\cite{wu2021contrastive}. Nevertheless, it is difficult for CNN-based models to learn long-range pixel dependencies.
    
     \noindent\textbf{Efficient Self-attention}. The computational complexity of the Transformer grows quadratically with the increasing spatial resolution of feature map, which makes it very demanding on computational resources. Some works reduce the computational burden by sliding window~\cite{ramachandran2019stand,hu2019local} or shifted window~\cite{liu2021swin,liang2021swinir,wang2022uformer,song2022vision} based self attention. However, this design limits the ability of Transformer to model long-distance dependencies in the data. MaxViT~\cite{tu2022maxvit} compensates for the decrease in receptive field with Grid attention. However, Grid attention is not strictly linear in computational complexity and is still quadratic on high-resolution images.
    Another approach is to modify the softmax-attention of the vanilla Transformer. Restormer~\cite{zamir2022restormer} applies self-attention between channels, and ignores global relationships between pixels. Performer~\cite{choromanski2020rethinking} achieves linear complexity by random projection. However, the queries, keys, and values require a large size, which results in increasing computation cost. Poly-nl~\cite{babiloni2021poly} bridges the connection between attention and high-order polynomials. However, this has not been explored in a self-attention structure.~\cite{katharopoulos2020transformers,shen2021efficient,qin2022cosformer,xu2021co,song2021ufo} decompose softmax by kernel functions and use the associative law of matrix multiplication to achieve linear complexity. These methods are functional approximations, e.g., each element in the attention map is positive~\cite{qin2022cosformer}. However, value approximations are not considered. Our method does not use the kernel function but performs the Taylor expansion on softmax directly, which guarantees a functional and numerical approximation to softmax. 
    
     \noindent\textbf{Multi-scale Transformer Networks}. In the field of high level vision,~\cite{mei2020pyramid} is a simple pyramid structure. IFormer~\cite{si2022inception} applies inception structures to mix high and low frequency information. However, it does not utilize different patch sizes. CrossViT~\cite{chen2021crossvit} and MPViT~\cite{lee2022mpvit} process multi-scale patches via multiple branches to obtain multi-scale receptive fields. However, the receptive field shape is not flexible. In the field of low level vision, MSP-Former~\cite{yang2022mstfdn} uses multi-scale projections to help Transformer to represent complex degraded environment. Giqe~\cite{shyam2022giqe} processes feature maps of different sizes via multi-branch. ~\cite{zhao2021complementary} represents different features related to the task utilizing multiple sub-network. The recent Transformer networks for recovery tasks~\cite{zamir2022restormer,wang2022uformer,song2022vision,ji2021u2,mei2020pyramid} build a simple U-net network or  with single-scale patches. However, these works hardly further explore multi-scale patches and multi-branch architectures.
     Although~\cite{kulkarni2023aerial} uses deformable convolution in self-attention, the number of sampling points of the convolution kernel is fixed. While our multi-scale deformable convolution not only has flexible sampling points, but also provides multi-level semantic information.

\section{MB-TaylorFormer}

    We aim to build an efficient and lightweight Transformer-based dehazing network. To reduce the computational complexity, we apply Taylor expansion of softmax-attention to satisfy the associative law and adopt a U-net structure similar to the Restormer~\cite{zamir2022restormer}. To compensate for the effects of Taylor expansion errors, we propose an MSAR module. In the following parts, we first describe the overall architecture of MB-TaylorFormer (Fig.~\ref{fig:pipeline}a). Then we introduce three core modules: multi-scale patch embedding (Fig.~\ref{fig:pipeline}b), Taylor expanded self-attention (Fig.~\ref{fig:pipeline}c) and MSAR module (Fig.~\ref{fig:pipeline}d). \\

\subsection{Multi-branch Backbone}
    Given a hazy image $ I\in \mathbb{R}^{3\times h\times w} $, we apply convolution for shallow feature extraction to generate $F_{\textbf{o}}\in \mathbb{R}^{c\times h\times w} $. Subsequently, we employ a four-stage encoder-decoder network for deep feature extraction. Each stage has a residual block containing a multi-scale patch embedding and a multi-branch Transformer block. We use multi-scale patch embedding to generate multi-scale tokens, and then feed them into multiple Transformer branches, respectively. Each Transformer branch contains multiple Transformer encoders. We apply the SKFF~\cite{zamir2020learning} module at the end of the multi-branch Transformer block to fuse features generated by different branches. Benefiting from the excellent performance of the Taylorformer and multi-branch design, we have the opportunity to compress the number of channels and Transformer blocks. We apply pixel-unshuffle and pixel-shuffle operations~\cite{shi2016real} at each stage to down-sample and up-sample features, respectively. Skip-connection~\cite{ronneberger2015u} is used to aggregate information from encoder and decoder, and a $ 1\times 1 $ convolutional layer is employed for dimensionality reduction (except for the first stage). We also use a residual block after the encoder-decoder structure to restore fine structural and textural details. Finally, a $ 3\times 3$ convolutional layer is applied to reduce channels and output a residual image $ R\in \mathbb{R}^{3\times h\times w}$. We thus obtain the restored
    image by $I'= I+R$. To further compress the computation and parameters, we apply depthwise separable convolutions~\cite{howard2017mobilenets,chollet2017xception} in the model.
\subsection{Multi-scale Patch Embedding}
\begin{figure}[t]
    \centering
    \includegraphics[width=7cm]{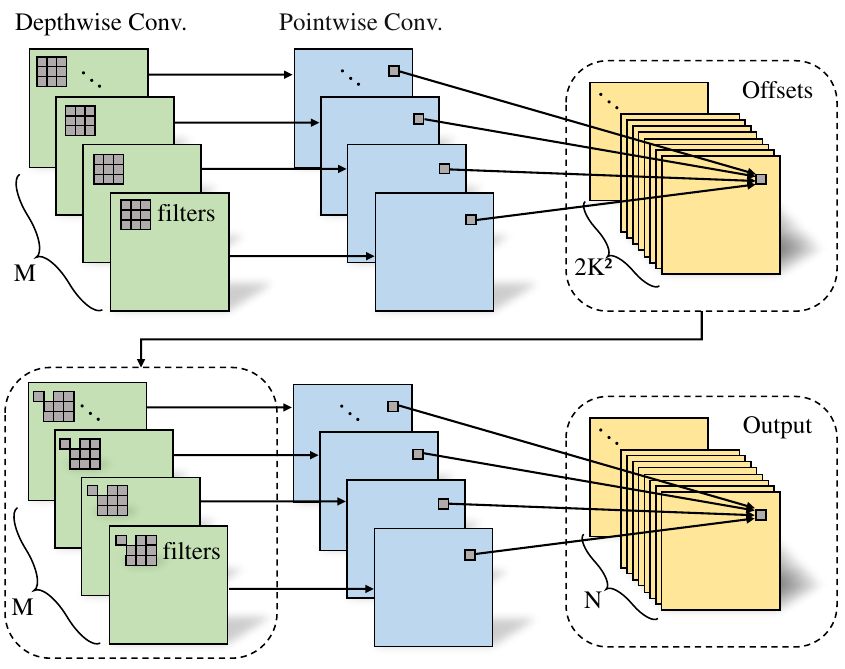}
    \caption{\textbf{An illustration of DSDCN.} The offsets are generated by $K \times K$ depthwise convolutions and pointwise convolutions, and the output is generated by $K \times K$ depthwise deformable convolutions and pointwise convolutions.}
    \label{fig:figure3}
    \vspace{-0.2cm}
\end{figure} 

    Visual elements can vary greatly in scale. Existing works~\cite{wang2022uformer,liang2021swinir,zamir2022restormer} employ convolution with fixed kernels into patch embedding, which may result in single scale of visual tokens. To address this issue, we design a new multi-scale patch embedding with three properties: 1) multiple sizes of receptive field, 2) multi-level semantic information, 3) flexible shapes of receptive field. Specifically, by designing multiple deformable convolutions (DCN)~\cite{dai2017deformable} with different scales of convolution kernels in parallel, we enable patch embedding to generate both coarse and fine visual tokens, as well as flexible transformation modeling capabilities. Inspired by the operation of stacking conventional layers that can expand receptive fields~\cite{simonyan2014very}, we stack several deformable convolutional layers with small kernels instead of a single deformable convolutional layer with large kernels. This not only increases the depth of network and consequently provides multi-level semantic information, but also helps to reduce parameters and computational burden. All deformable convolutional layers are followed by Hardswish~\cite{howard2019searching} activation functions.

    Similar to the strategy of depthwise separable convolutions~\cite{howard2017mobilenets,chollet2017xception}, we propose depthwise separable and deformable convolutions (DSDCN), which decomposes the parts of DCN with depthwise convolution and pointwise convolution, as shown in Fig.~\ref{fig:figure3}. The computational cost of standard DCN and DSDCN regarding a $h\times w$ image is as follows: 
\begin{equation}    
 \Omega(\mathrm{DCN})=2M K^{4} h w+M N K^{2} h w +4M K^{2} h w,
\end{equation}
 \vspace{-0.5cm} 
 \begin{equation}
 \begin{split}
 \Omega(\mathrm{DSDCN})=8M K^{2} h w+M N h w.
  \end{split}
\end{equation}
    where M and N are the numbers of channels in the input and output, respectively, and K is the kernel size of convolution. The number of parameters of DCN and DSDCN are as follows:
\begin{equation}    
\mathrm{P}(\mathrm{DCN})=2M K^{4}+M N K^{2},
\end{equation}
 \vspace{-0.5cm} 
 \begin{equation}    
 \mathrm{P}(\mathrm{DSDCN})=4 M K^{2}+M N,
\end{equation}

    \begin{figure}[t]
    \centering
    \includegraphics[width=8cm]{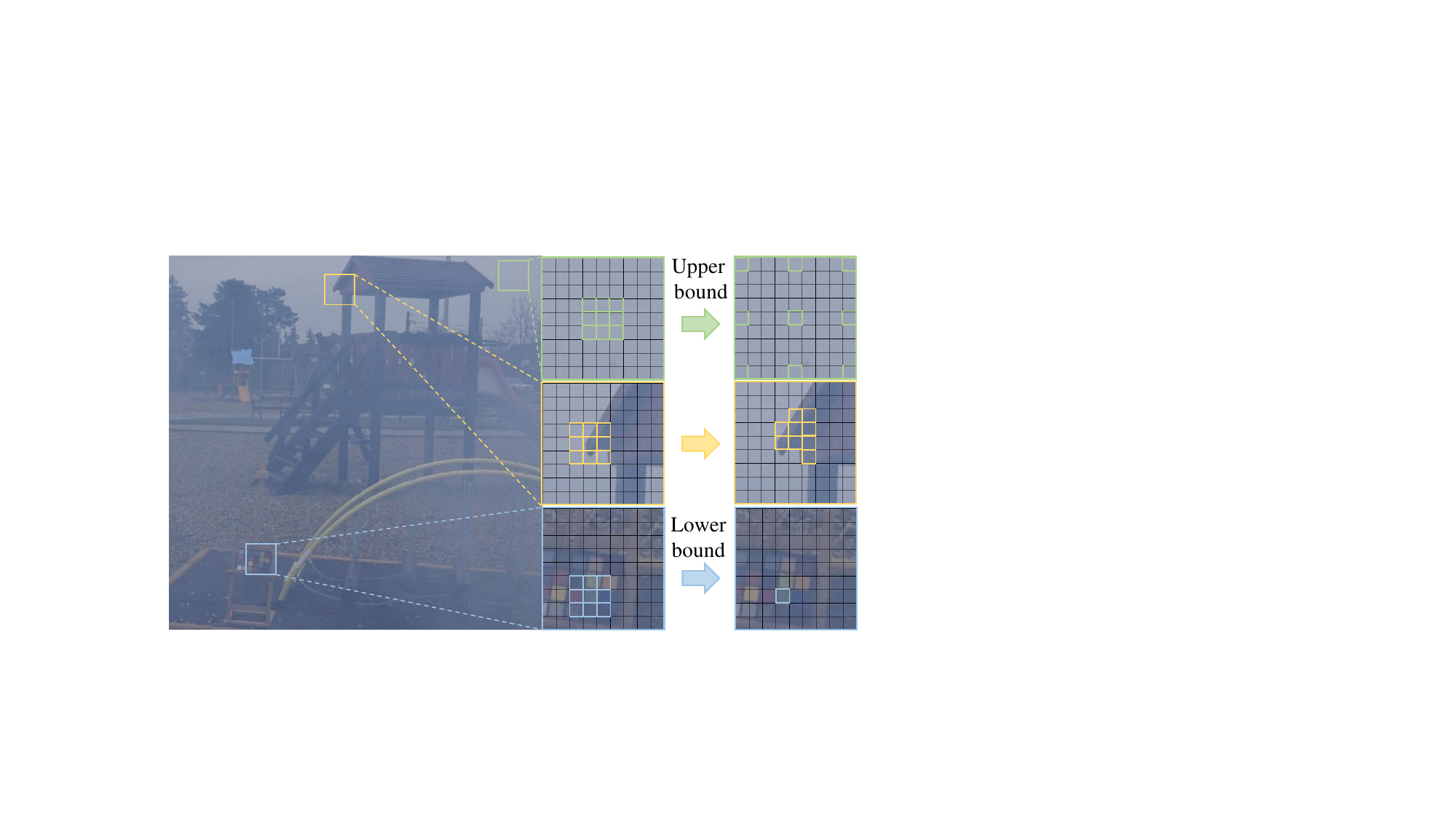}
    \caption{\textbf{An illustration of the receptive field of DSDCN (the offsets are truncated to [-3,3]).} The upper bound of the receptive field of the DSDCN is $9\times 9$ and the lower bound is $1\times 1$.}
    \label{fig:figure4}
    \vspace{-0.2cm}
\end{figure} 
In summary, DSDCN greatly reduces computational complexity and number of parameters compared to DCN.   
    
Considering that images are locally relevant and patch embedding captures the basic elements of feature maps, the visual elements (\textit{i.e.}, tokens) should be more focused on local areas. We control the receptive field range of the patch embedding layer by truncating the offsets, which we choose in practice to be $[-3,3]$. As it is shown in Fig.~\ref{fig:figure4}, depending on the shape of visual object, the model is able to select the receptive field size by learning, which has an upper bound of $ 9\times 9 $, equivalent to a dilated convolution~\cite{yu2015multi} of BF = 4, and a lower bound of $ 1\times 1 $. When we set up multi-scale patch embedding in parallel, the sizes of the receptive field of different branches are $ x\in [1,9] $, $ y\in [x,x+8] $ and $ z\in [y,y+8] $ in ascending order (for three branches). Experiments in the supplementary material show that setting constraints on the receptive field of each token in a reasonable way can improve the model's performance.

\subsection{Taylor Expanded Multi-head Self-Attention}

    Let queries ($Q$), keys ($K$), and values ($V$) be a sequence of $h \times w$ feature vectors with dimensions $D$, where $h$ and $w$ is the height and width of the image, respectively. The formula of the vanilla Transformer~\cite{vaswani2017attention} is as follows:
\begin{equation}
V' = Softmax\left(\frac{Q^{T}K}{\sqrt{D}}\right)V^{T}.\label{con:att}
\end{equation}
    Since $Q\in \mathbb{R}^{hw\times D}$, $K\in \mathbb{R}^{hw\times D}$ and $V\in \mathbb{R}^{hw\times D}$ , Softmax causes the computational complexity of self-attention to be $\mathcal{O}\left(h^{2}w^{2}\right)$, resulting in expensive computational cost.
    
\begin{figure}[t]
    \centering
    \includegraphics[width=6cm]{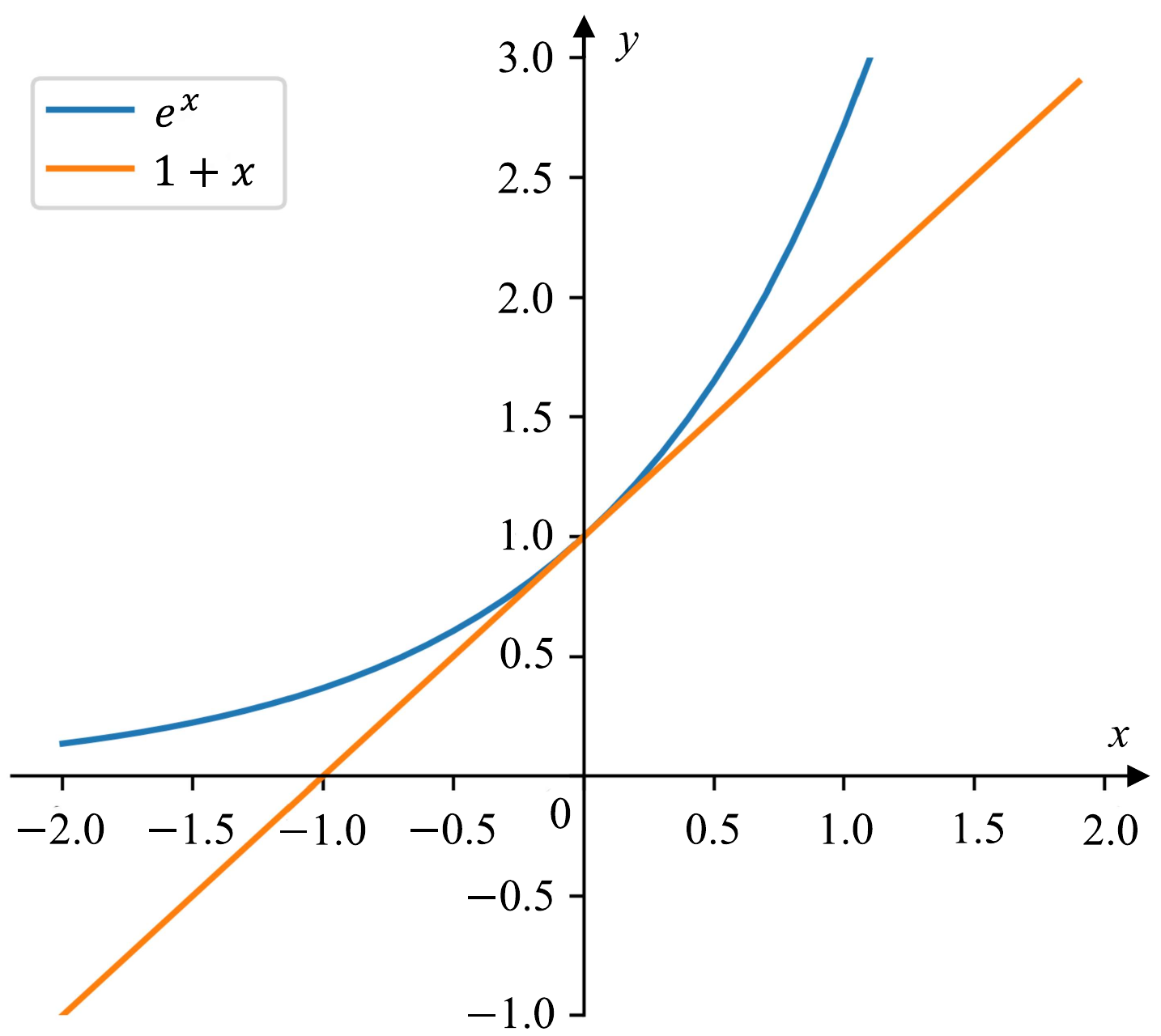}

    \caption{\textbf{$\mathbf{e^{x}}$ (blue) and its first-order Taylor expansion curve (orange).} The closer to 0, the tighter the approximation of the orange line to the blue line.}
    \label{fig:figure5}
    \vspace{-0.2cm}
\end{figure}

    We aim to reduce the computational complexity of self-attention from $\mathcal{O}\left(h^{2}w^{2}\right)$ to $\mathcal{O}\left(hw\right)$. To achieve it, we first write the generalized attention equation for Eq.~\eqref{con:att}, as follows:
\begin{equation}    
V_{i}^{\prime}=\frac{\sum_{j=1}^{N} f\left(Q_{i}, K_{j}\right) V_{j}}{\sum_{j=1}^{N} f\left(Q_{i}, K_{j}\right)},\label{con:att2}
\end{equation}
where the matrix with $i$ as subscript is the vector of the $i$-th row of matrix, and $f(\cdot)$ denotes any similarity function. Eq.~\eqref{con:att2} degenerates to Eq.~\eqref{con:att} when we let $f\left(Q_{i}, K_{j}\right)=\exp \left(\frac{Q_{i}^{T} K_{j}}{\sqrt{D}}\right)$.
    If we apply the Taylor formula to perform a first-order Taylor expansion on $\exp \left(\frac{Q_{i}^{T} K_{j}}{\sqrt{D}}\right)$ at 0, we can rewrite Eq.~\eqref{con:att2} as:
\begin{equation}    
V_{i}^{\prime}=\frac{\sum_{j=1}^{N} \left(1+Q_{i}^{T}K_{j}+o\left(Q_{i}^{T}K_{j}\right) \right)V_{j}^{T}}{\sum_{j=1}^{N} \left(1+Q_{i}^{T}K_{j}+o\left(Q_{i}^{T}K_{j}\right)\right)}.\label{con:att3}
\end{equation}
Further, we generate $\tilde{Q_{i}}$, $\tilde{K_{i}}$ from the normalization of vectors $Q_{i}$ and $K_{j}$ to approximate $\exp \left(\frac{Q_{i}^{T} K_{j}}{\sqrt{D}}\right)$. When the norm of $\tilde{Q_{i}}$ and $\tilde{K_{i}}$ are smaller than $1$, we can make the values of the attention map all positive, and in practice, we find the best results are achieved by normalizing the norm to $0.5$. As shown in Fig.~\ref{fig:figure5}, we consider that there is an approximation to $e^{x}$ and its first-order Taylor expansion is in the definition domain of $[-0.25,0.25]$. So we eliminate Peano's form of remainder~\cite{peano1889nouvelle} and obtain the expression for the Taylor expansion of self-attention as follows:
\begin{equation} 
V_{i}^{\prime}=\frac{\sum_{j=1}^{N} \left(1+\tilde{Q}_{i}^{T}\tilde{K}_{j} \right)V_{j}^{T}}{\sum_{j=1}^{N} \left(1+\tilde{Q}_{i}^{T}\tilde{K}_{j}\right)}.\label{con:att4}
\end{equation}
Finally, we apply the matrix multiplication associative law to Eq.~\eqref{con:att4}, as follows:
\begin{equation}   
\begin{split}
V_{i}^{\prime}=\text {Taylor-Attention}\left(Q_{i},K_{i},V_{i}\right)\\
=\frac{\sum_{j=1}^{N} V_{j}^{T}+\tilde{Q_{i}}^{T}\sum_{j=1}^{N} \tilde{K}_{j}V_{j}^{T}}{N+\tilde{Q}_{i}^{T}\sum_{j=1}^{N} \tilde{K}_{j}}.\label{con:att5}
\end{split}
\end{equation}
Supplementary materials provide the pseudo-code of Taylor expanded multi-head self-attention (T-MSA).

Similar to MDTA~\cite{zamir2022restormer}, we employ deep convolutional generation for $Q$, $K$, $V$ to emphasize the local context. We also employ a multi-head structure, where the number of heads increases progressively from top to bottom of the level. The computational complexity of the standard multi-head self-attention (MSA) module and the T-MSA module regarding an image of $h \times w$ patches is as follows:
 \begin{equation}    
  \Omega(\mathrm{MSA})=4 hw D^{2}+2h^{2}w^{2} D,
\end{equation}
 \vspace{-0.5cm} 
 \begin{equation}    
  \Omega(\mathrm{T} \mbox{-} \mathrm{MSA})=18hwD + 7hw D^{2}.
\end{equation}
Generally, $h \times w$ is much larger than $D$, so T-MSA provides more possibilities than MSA for testing high-resolution images, and ensures the values are close to the MSA.

\begin{table*}[t]
 \centering
\caption{\textbf{Quantitative comparisons of various methods on dehazing benchmarks.}``-" indicates that the result is not available. The best and second best results are highlighted in bold and underlined, respectively.}
    \label{tab:compare}
    \scalebox{0.88}{
\centering
\begin{tabular}{c|cc|cc|cc|cc|cc}
 \toprule
\multirow{2}{*}{Methods}                                & \multicolumn{2}{c|}{SOTS-Indoor} & \multicolumn{2}{c|}{SOTS-Outdoor} & \multicolumn{2}{c|}{O-HAZE}                                              & \multicolumn{2}{c|}{Dense-Haze}                                          & \multicolumn{2}{c}{Overhead}          \\ \cline{2-11} 
                                                        & PSNR$\uparrow$            & SSIM$\uparrow$            & PSNR$\uparrow$             & SSIM$\uparrow$            & PSNR$\uparrow$                               & SSIM$\uparrow$                                & PSNR$\uparrow$                                & SSIM$\uparrow$                                & \multicolumn{1}{c|}{\#Param} & MACs    \\ \hline
DCP~\cite{he2010single}           & 16.62           & 0.818          & 19.13           & 0.815           & \multicolumn{1}{c}{16.78}          & \multicolumn{1}{c|}{0.653}          & \multicolumn{1}{c}{12.72}          & \multicolumn{1}{c|}{0.442}          & \multicolumn{1}{c|}{-}       & 0.6G \\
DehazeNet~\cite{cai2016dehazenet}                                               & 19.82           & 0.821          & 24.75           & 0.927           & 17.57                              & \multicolumn{1}{c|}{0.770}           & 13.84                              & 0.430                                & \multicolumn{1}{c|}{0.01M}  & 0.6G \\
GFN~\cite{ren2018gated}                                                     & 22.30           & 0.880          & 21.55           & 0.844           & 18.16                              & 0.671                               & -                                  & -                                   & \multicolumn{1}{c|}{0.50M}  & 14.9G \\
GDN~\cite{liu2019griddehazenet}    & 32.16           & 0.984          & 30.86           & 0.982           & \multicolumn{1}{c}{18.92}          & \multicolumn{1}{c|}{0.672}          & \multicolumn{1}{c}{14.96}          & 0.536                               & \multicolumn{1}{c|}{0.96M}  & 21.5G \\
PFDN~\cite{dong2020physics}                                                    & 32.68           & 0.976          & -               & -               & -                                  & -                                   & -                                  & -                                   & \multicolumn{1}{c|}{11.27M}  & 51.5G \\
MSBDN~\cite{dong2020multi}                     & 33.67           & 0.985          & 33.48           & 0.982           & 24.36                              & 0.749                               & \multicolumn{1}{c}{15.13}          & 0.555                               & \multicolumn{1}{c|}{31.35M}  & 41.54G \\
FFA-Net~\cite{qin2020ffa}         & 36.39           & 0.989          & 33.57           & 0.984           & \multicolumn{1}{c}{22.12}          & \multicolumn{1}{c|}{0.770}           & \multicolumn{1}{c}{15.70}          & 0.549                               & \multicolumn{1}{c|}{4.46M}  & 287.8G \\
AECR-Net~\cite{wu2021contrastive} & 37.17           & 0.990          & -               & -               & -                                  & -                                   & 15.80                              & 0.466                               & \multicolumn{1}{c|}{2.61M}  & 52.2G \\
MAXIM-2S~\cite{tu2022maxim}       & 38.11           & 0.991          & 34.19           & 0.985           & -                                  & -                                   & -                                  & -                                   & \multicolumn{1}{c|}{14.10M}  & 216.0G \\
SGID-PFF~\cite{bai2022self}                  & 38.52           & 0.991          & 30.20           & 0.975           & 20.96                                  & 0.741                                  & 12.49                                  & 0.517                                   & \multicolumn{1}{c|}{13.87M}  & 156.4G \\Restormer~\cite{zamir2022restormer}                   & 38.88           & 0.991          & -           & -           & 23.58                             & 0.768                               & 15.78                       & 0.548                      & \multicolumn{1}{c|}{26.10M}  & 141.0G \\   
Dehamer~\cite{guo2022image}                   & 36.63           & 0.988          & 35.18           & 0.986           & \underline{25.11}                             & 0.777                               & {\underline{16.62} }                        & \underline{0.560}                      & \multicolumn{1}{c|}{132.50M}  & 60.3G \\ \hline
Ours (-B)                                          & \underline{40.71}    & \underline{0.992}    & \underline{37.42}              & \underline{0.989}               & 25.05                                  & \textbf{0.788}                                   & \multicolumn{1}{c}{\textbf{16.66}} & \multicolumn{1}{c|}{\underline{0.560}} & \multicolumn{1}{c|}{2.68M}  & 38.5G \\
Ours (-L)                                        & \textbf{42.64}  & \textbf{0.994} & \textbf{38.09}             & \textbf{0.991}              & \multicolumn{1}{c}{\textbf{25.31}} & \multicolumn{1}{c|}{\underline{0.782}} & 16.44                                 & \textbf{0.566}                                  & \multicolumn{1}{c|}{7.43M}  & 88.1G \\ \bottomrule
\end{tabular}}
\vspace{-0.2cm}
\end{table*}

    \subsection{Multi-scale Attention Refinement}
    Since we perform a first-order Taylor expansion of softmax in T-MSA and ignore Peano's form of reminder, there is an inevitable approximation error. For the $n$-th order remainder term  $\frac{(Q_{i}K_{j})^{n}}{n!}  (n\ge 2 )$ of Taylor expansion, the matrix multiplicative combination law cannot be used to make the computational complexity of T-MSA linear. However, the remainder term is related to the Q and K matrices. Considering that the images have local correlation, we learn the local information of the Q and K matrices to correct the inaccurate output $V'$. In addition, the conv-attention module allows TaylorFormer to better handle high frequency information~\cite{park2022vision}. 
    
    Specifically, for multi-heads $Q_{m}\in \mathbb{R}^{head\times \frac{D}{head} \times N }$ and $K_{m}\in \mathbb{R}^{head\times \frac{D}{head} \times N }$ , we reshape them into $\hat{Q}_{m}\in \mathbb{R}^{ head \times \frac{D}{head}\times H\times W }$  and $\hat{K }_{m}\in \mathbb{R}^{ head \times \frac{D}{head}\times H\times W  }$, where $head$ denotes the number of heads, and we concatenate $\hat{Q}_{m}$ and $\hat{K}_{m}$ along the channel dimension to generate a tensor $ T\in \mathbb{R}^{ head \times \frac{2D}{head}\times H\times W}$, which is subsequently passed through a multi-scale grouped convolutional layer to generate a
   gating tensor $ G\in \mathbb{R}^{ head \times 1\times H\times W  }$ as:
     \begin{equation}   
G=\text {Sigmoid}\left( \text {Concat}\left( T_{1} W_{1}^{G}, \ldots,  T_{head} W_{head}^{G}\right)\right),
\end{equation}
where $T_{head}\in \mathbb{R}^{\frac{D}{head} \times H \times W } $ is the $head$-th head of $T$, and $W_{(\cdot)}^{G}$ is the convolution with different kernels. Since different levels of the network have different numbers of heads, we choose the corresponding multi-scale grouped convolutions for different numbers of heads. Supplementary materials provide details of the structure of the MSAR module.

With the approach of T-MSA and the module of MSAR, the refined T-MSA module is computed as:
\begin{equation}   
\begin{split}
\hat{X} =X+\text {Cat}\left( H_{1}\odot  G_{1}, \ldots,  H_{head} \odot G_{head}\right)W^{p},\\
H_{i}  =\text {Taylor-Attention}\left(QW_{i}^{Q},KW_{i}^{K},VW_{i}^{V}\right),
\end{split}
\end{equation}
where $X$ and $\hat{X}$ denote the input and output feature maps. The projections are parameter matrices $W^{P}\in \mathbb{R}^{D\times D }$, $W_{i}^{Q}\in \mathbb{R}^{D\times\frac{D}{head} }$, $W_{i}^{K}\in \mathbb{R}^{D\times\frac{D}{head} }$, and $W_{i}^{V}\in \mathbb{R}^{D\times\frac{D}{head} }$.

\section{Experiments}

In this section, we conduct experiments to demonstrate the effectiveness of the proposed MB-TaylorFormer. Further details including more qualitative results are provided in Supplementary Materials.

\subsection{Experiment Setup}
\textbf{Implementation Details}. We provide two architectures of MB-TaylorFormer including MB-TaylorFormer-B (the basic model) and MB-TaylorFormer-L (a larger variant). 
%
We employ random cropping and random flipping for data augmentation. We set the initial learning rate to 2e-4 and gradually reduce it to 1e-6 using cosine annealing~\cite{loshchilov2016sgdr}. we only use L1 loss as our loss function.

\textbf{Datasets}. We evaluate the proposed MB-TaylorFormer on synthetic dataset (RESIDE~\cite{8451944}) and real-world datasets (O-HAZE~\cite{ancuti2018haze}, Dense-Haze~\cite{ancuti2019dense}). As subsets of RESIDE, ITS and OTS contain $13990$ pairs of indoor and $313950$ pairs of outdoor images, respectively. Models are evaluated on the SOTS subset. 
The real-world datasets, O-HAZE and Dense-Haze, contain $45$ and $55$ paired images, respectively. We use the last $5$ images of each dataset as the testing set and the rest as the training set.

\begin{figure*}[t]
  \captionsetup[subfloat]{labelformat=empty}
  \captionsetup{font={small,stretch=0.5}, justification=raggedright}
  	\begin{minipage}[b]{0.7\columnwidth}
		\centering
		\subfloat[Input (7.27/0.616)
  ]{
    \includegraphics[width=1\columnwidth]{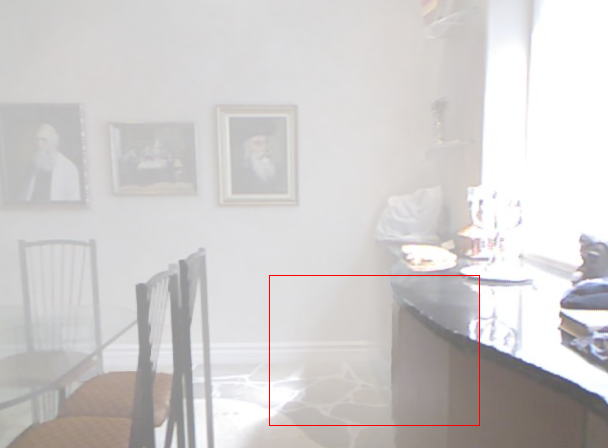}
}
	\end{minipage}
	\begin{minipage}[b]{0.32\columnwidth}
		\centering
		\subfloat[GDN (25.19/0.954)
  ]{
    \includegraphics[width=1\columnwidth]{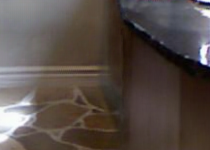}
  }
  \subfloat[MSBDN (25.65/0.961)
  ]{
    \includegraphics[width=1\columnwidth]{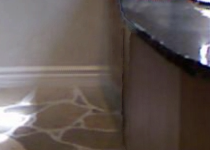}
  }
  \subfloat[FFA-Net (30.70/0.966)
  ]{
    \includegraphics[width=1\columnwidth]{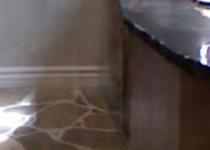}
  }
  \subfloat[MAXIM (30.20/0.981)
  ]{
    \includegraphics[width=1\columnwidth]{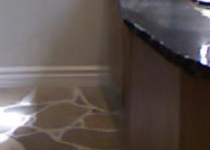}
  }		
		\\
	\subfloat[SGID-PFF (30.85/0.977)
  ]{
    \includegraphics[width=1\columnwidth]{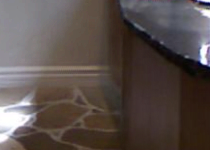}
  }	
   \subfloat[Dehamer (33.28/0.979)
  ]{
    \includegraphics[width=1\columnwidth]{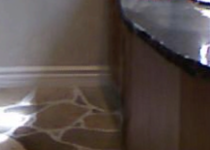}
  }
  		\subfloat[Ours (37.71/0.988)
  ]{
    \includegraphics[width=1\columnwidth]{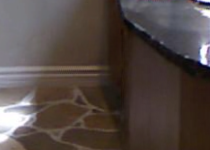}
  }
    		\subfloat[GT (PSNR(dB)/SSIM)
  ]{
    \includegraphics[width=1\columnwidth]{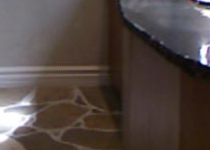}
  }
	\end{minipage}\\
 \begin{minipage}[b]{0.7\columnwidth}
		\centering
		\subfloat[Input (15.20/0.893)
  ]{
    \includegraphics[width=1\columnwidth]{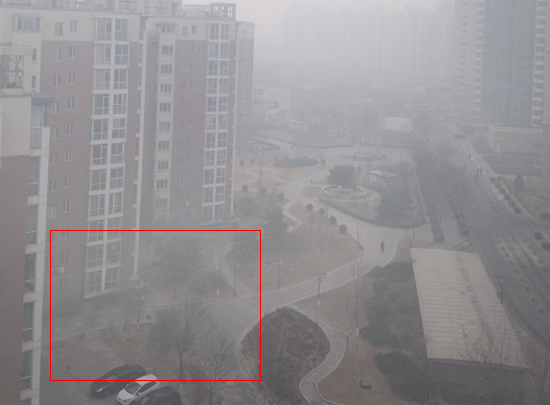}
  }

	\end{minipage}
	\begin{minipage}[b]{0.32\columnwidth}
		\centering
		\subfloat[GDN (25.55/0.977)
  ]{
    \includegraphics[width=1\columnwidth]{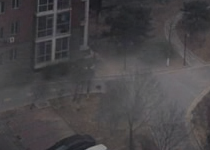}
  }
  \subfloat[MSBDN (30.08/0.981)
  ]{
    \includegraphics[width=1\columnwidth]{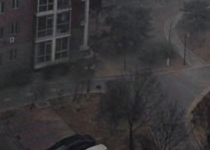}
  }
  \subfloat[FFA-Net (24.74/0.977)
  ]{
    \includegraphics[width=1\columnwidth]{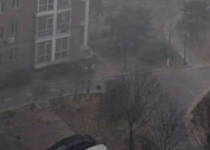}
  }
  		\subfloat[MAXIM (27.91/0.984)
  ]{
    \includegraphics[width=1\columnwidth]{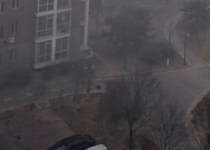}
  }
		\\
 \subfloat[SGID-PFF (19.60/0.924)
  ]{
    \includegraphics[width=1\columnwidth]{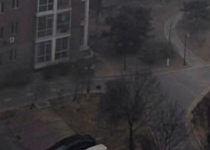}
  }
  		\subfloat[Dehamer (34.60/0.991)
  ]{
    \includegraphics[width=1\columnwidth]{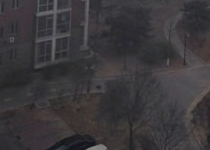}
  }
  		\subfloat[Ours (36.34/0.991)
  ]{
    \includegraphics[width=1\columnwidth]{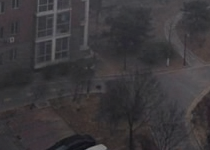}
  }
    		\subfloat[GT (PSNR(dB)/SSIM)
  ]{
    \includegraphics[width=1\columnwidth]{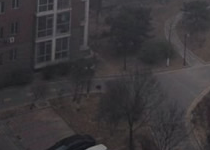}
  }
	\end{minipage}
  \caption{\textbf{Visual comparisons on synthetic hazy images.} The first two rows of images are from the SOTS-Indoor dataset, the last two rows are from SOTS-Outdoor. Our MB-TaylorFormer-L generates haze-free images with color fidelity and finer textures. }
    \label{fig:comparision}

\end{figure*}

\begin{figure*}[t]
  \centering

  \captionsetup[subfloat]{labelformat=empty}
    \subfloat[Input 
  ]{
    \includegraphics[width=0.293\columnwidth]{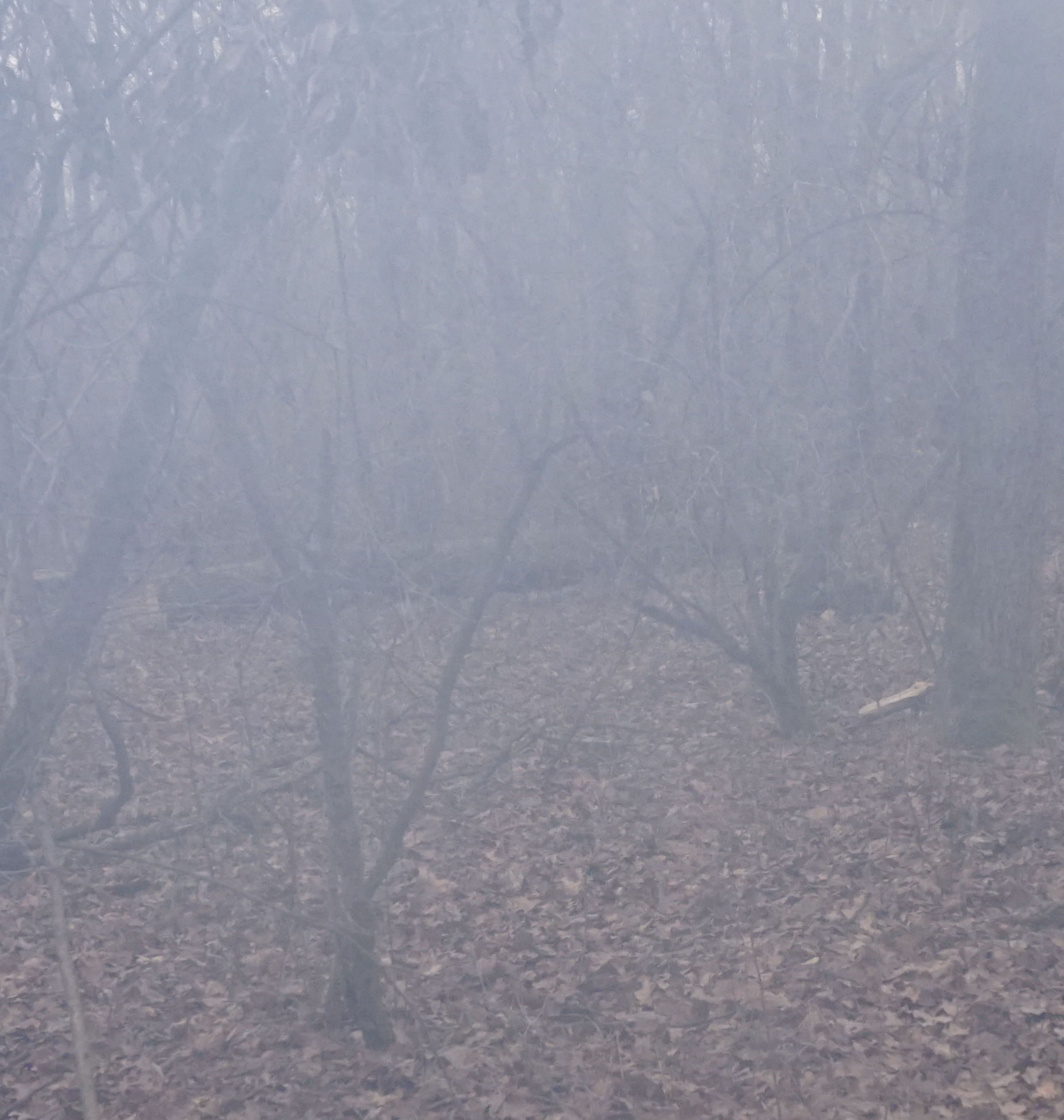}
  }
  \subfloat[MSBDN
  ]{
    \hspace{-0.1in}
    \includegraphics[width=0.293\columnwidth]{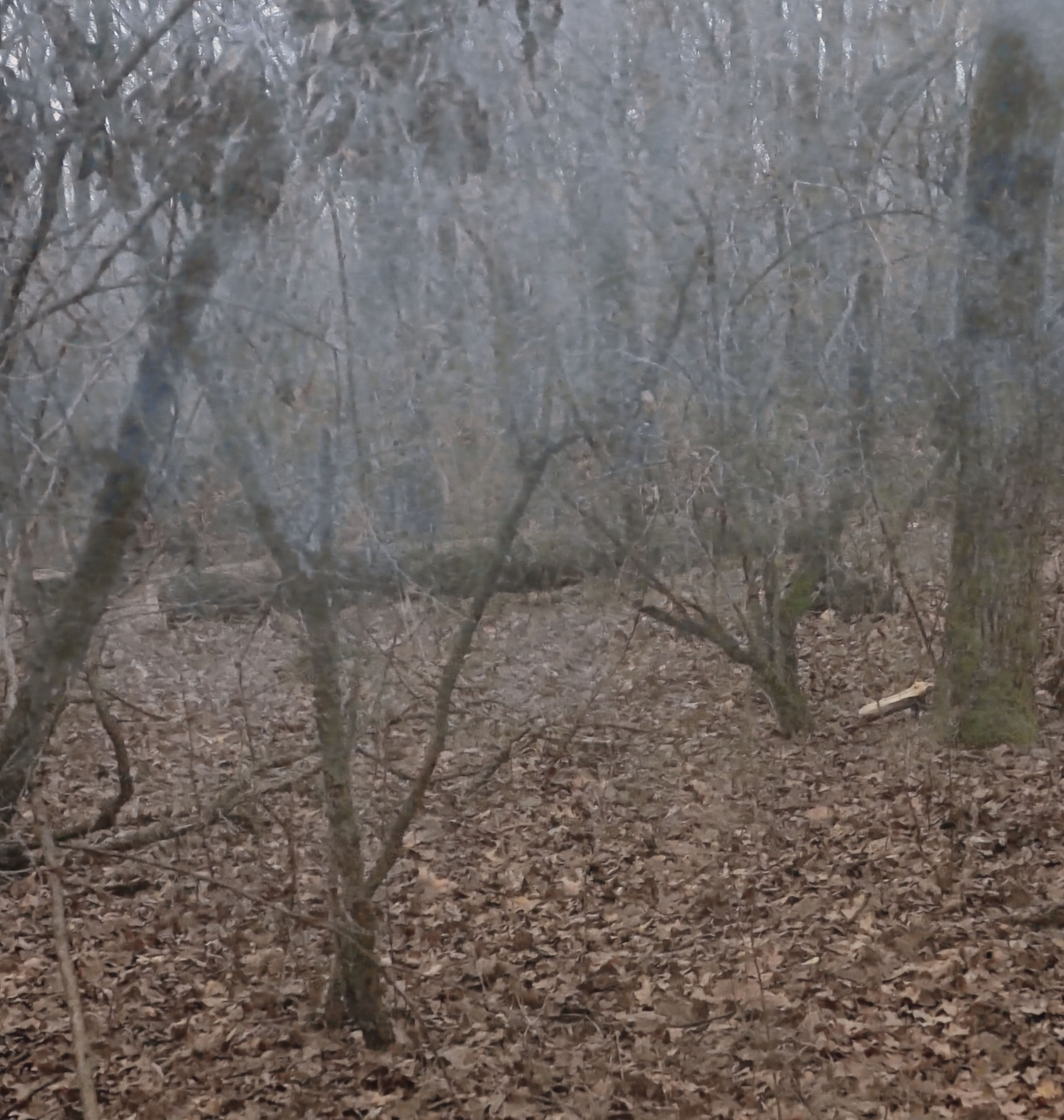}
  }
  \subfloat[FFA-Net 
  ]{
    \hspace{-0.1in}
    \includegraphics[width=0.293\columnwidth]{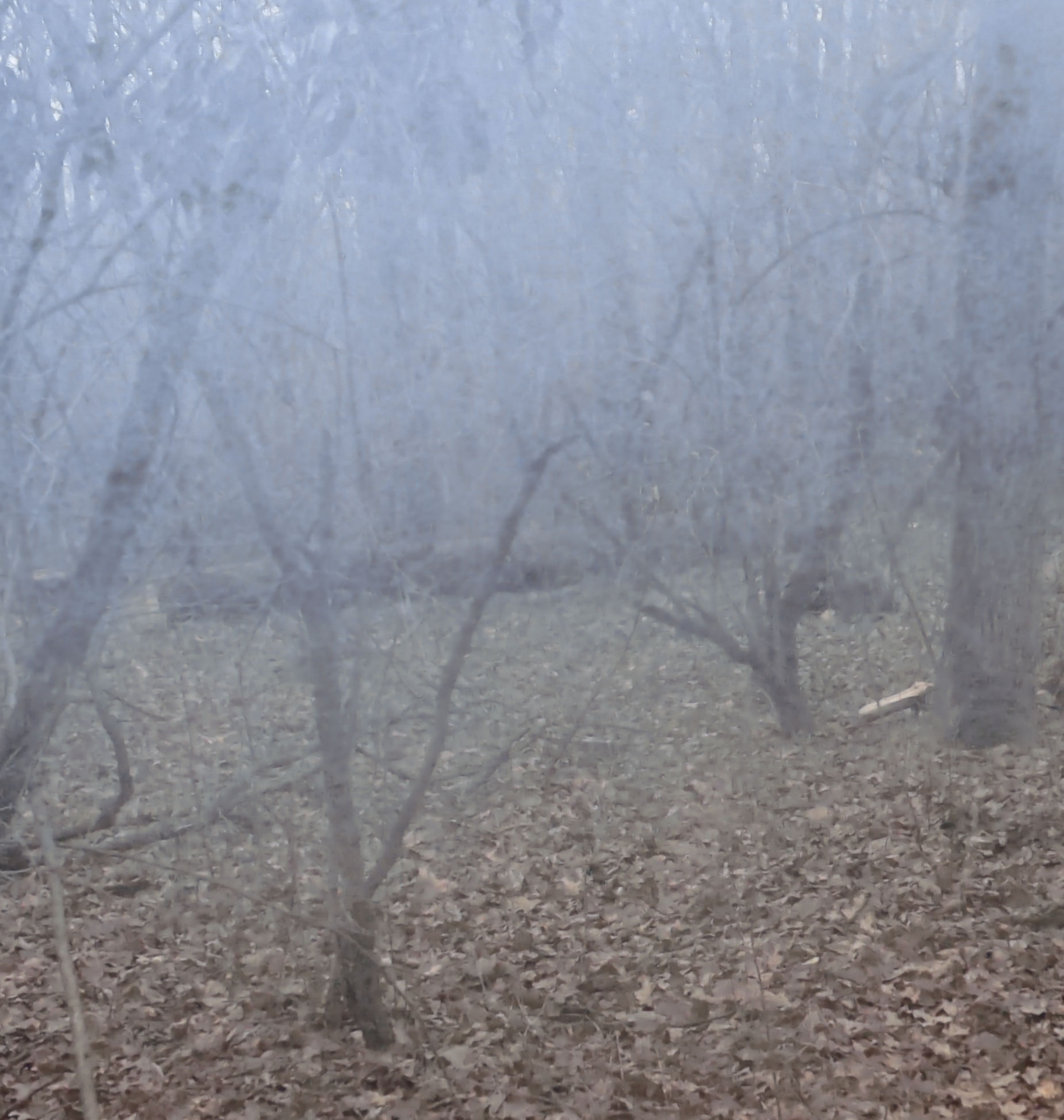}
  }
    \subfloat[SGID-PFF
  ]{
    \hspace{-0.1in}
    \includegraphics[width=0.293\columnwidth]{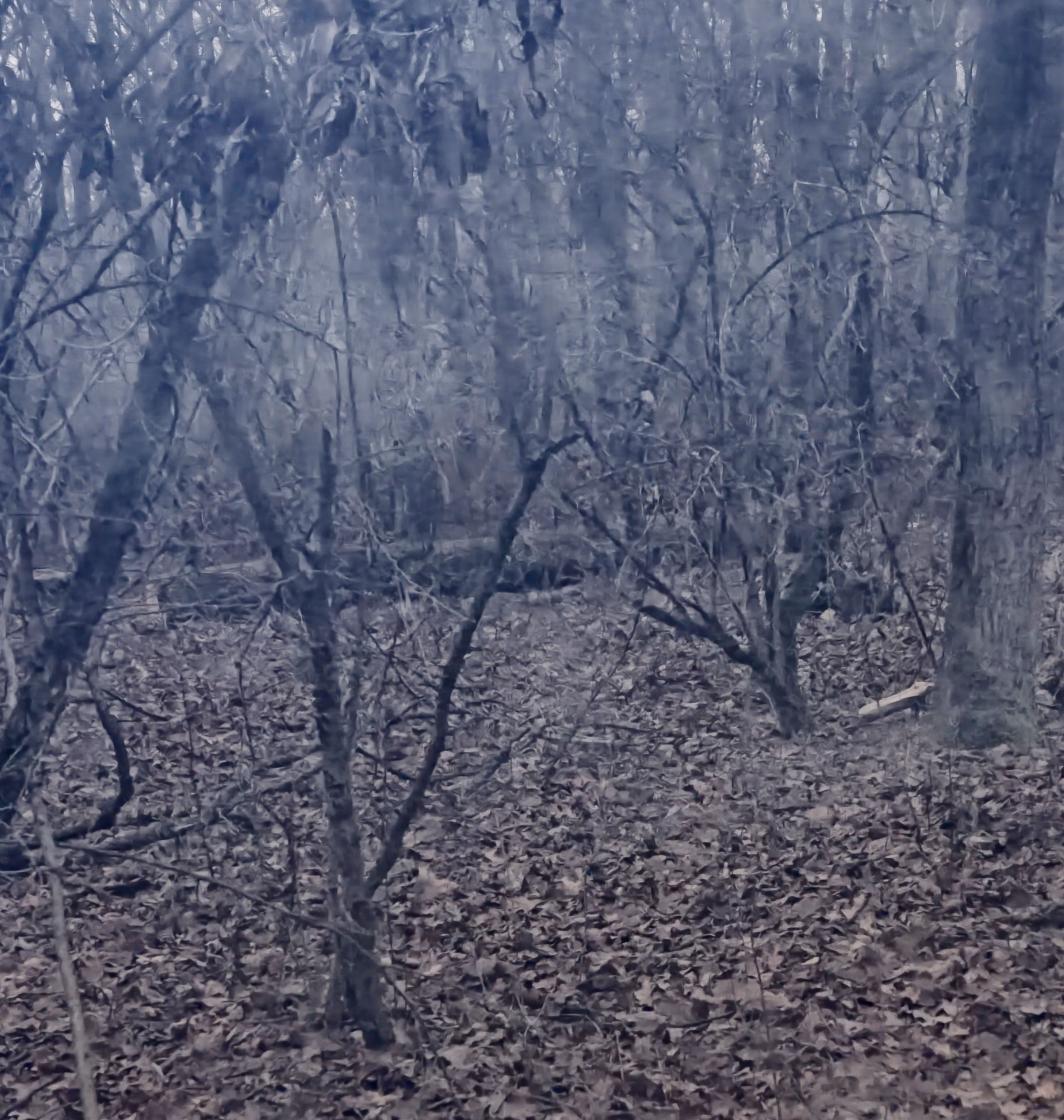}
  }
  \subfloat[Dehamer
  ]{
   \hspace{-0.1in}
    \includegraphics[width=0.293\columnwidth]{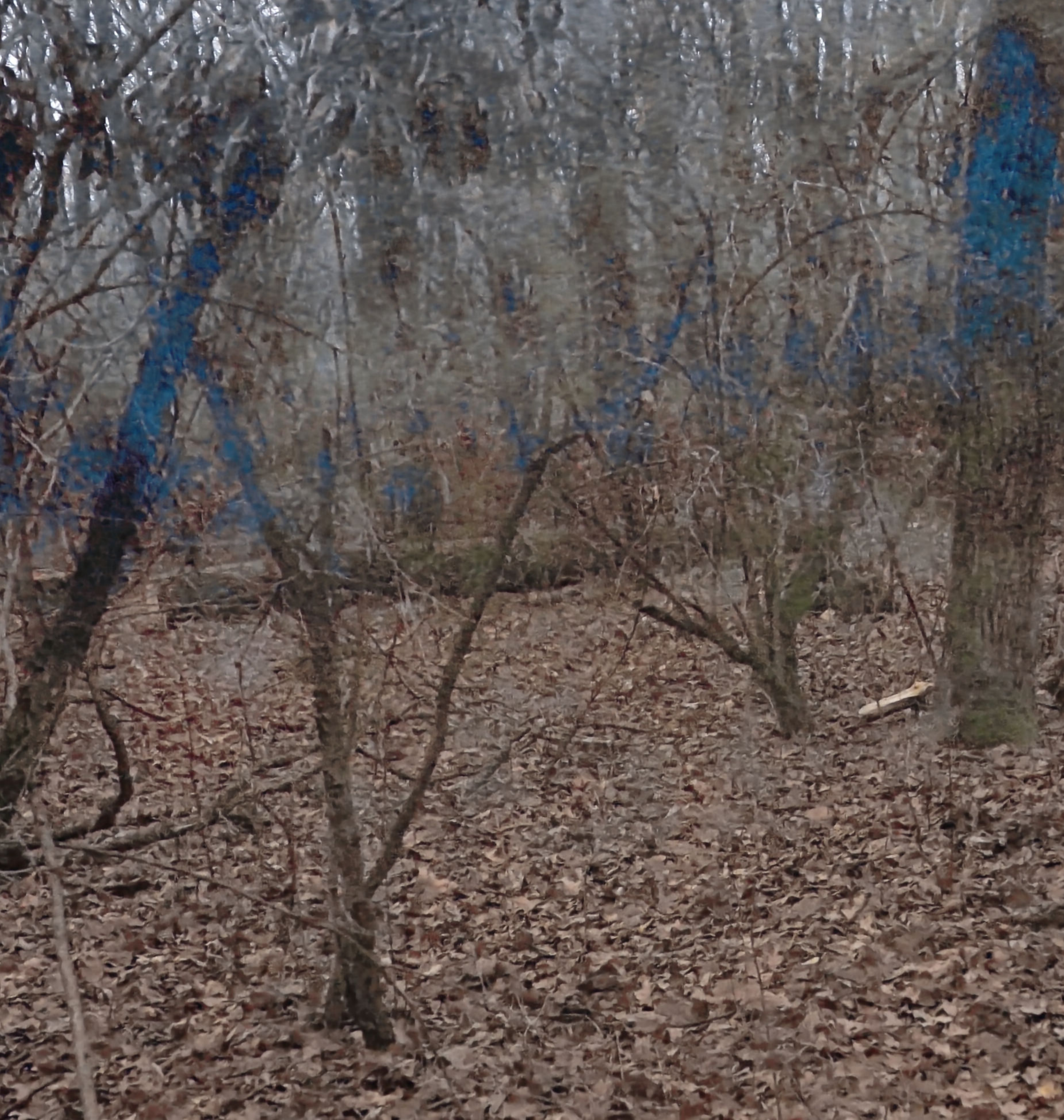}
  }
  \subfloat[Ours
  ]{
    \hspace{-0.1in}
    \includegraphics[width=0.293\columnwidth]{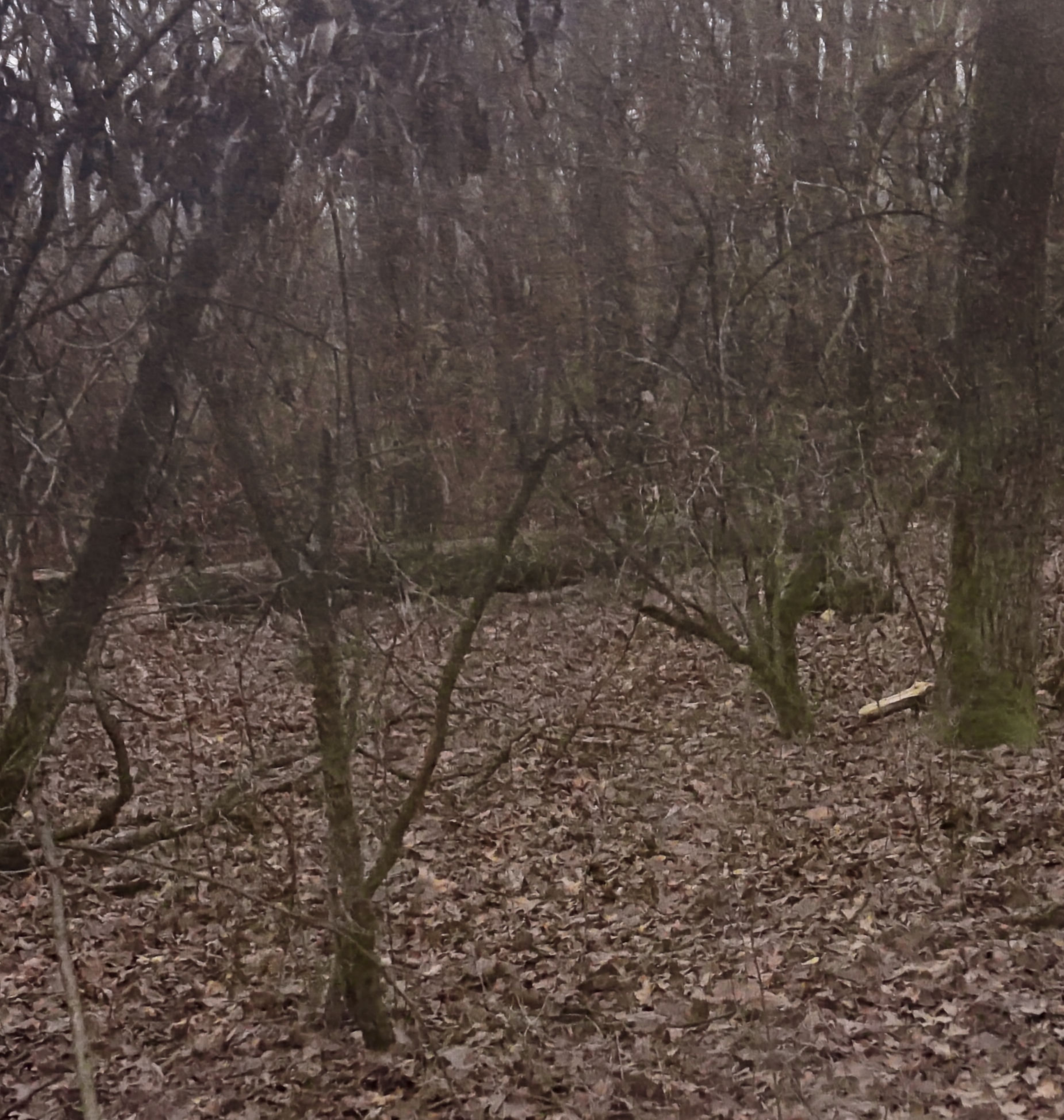}
  }
  \subfloat[GT
  ]{
    \hspace{-0.1in}
    \includegraphics[width=0.293\columnwidth]{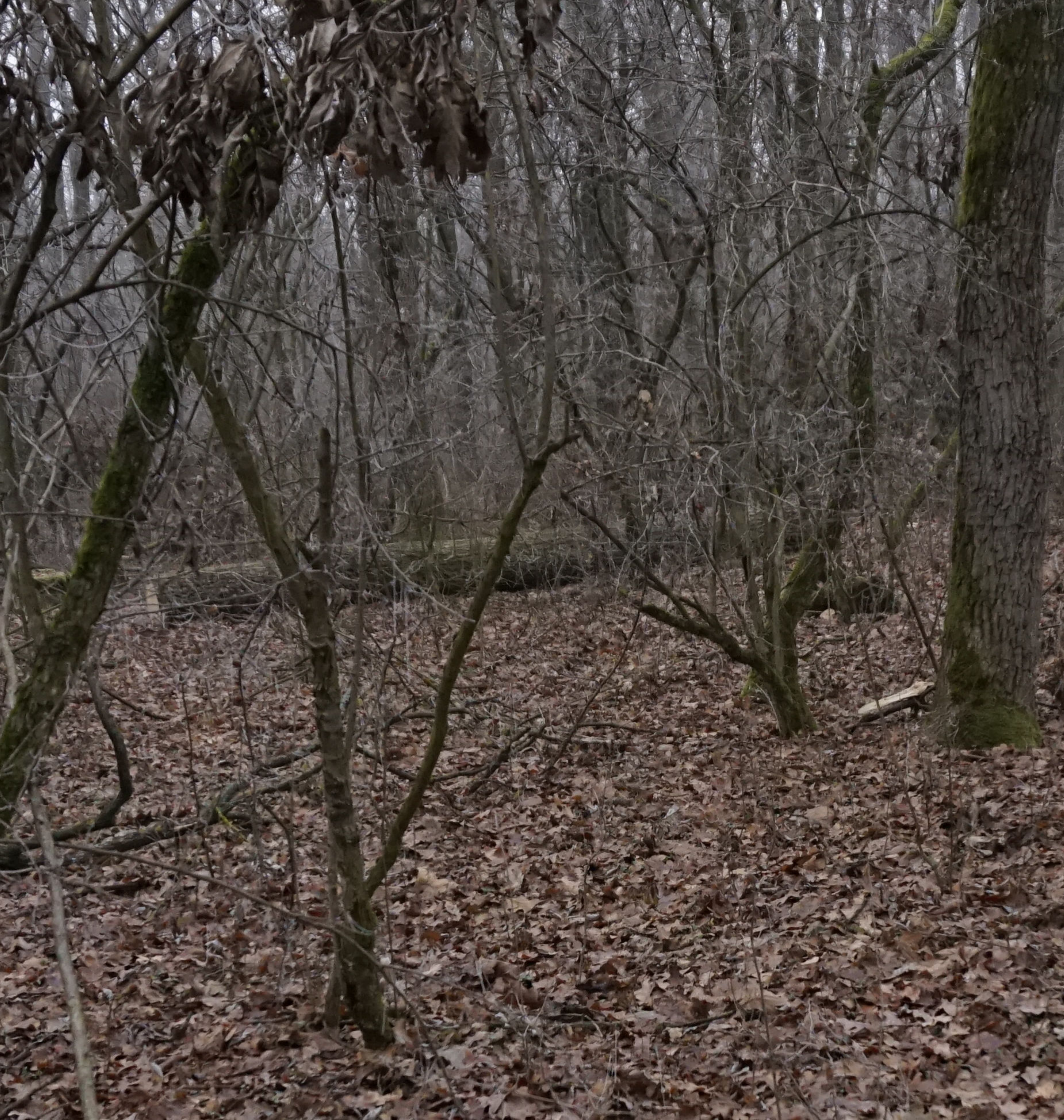}
  }\\
  \subfloat[Input 
  ]{
    \includegraphics[width=0.293\columnwidth]{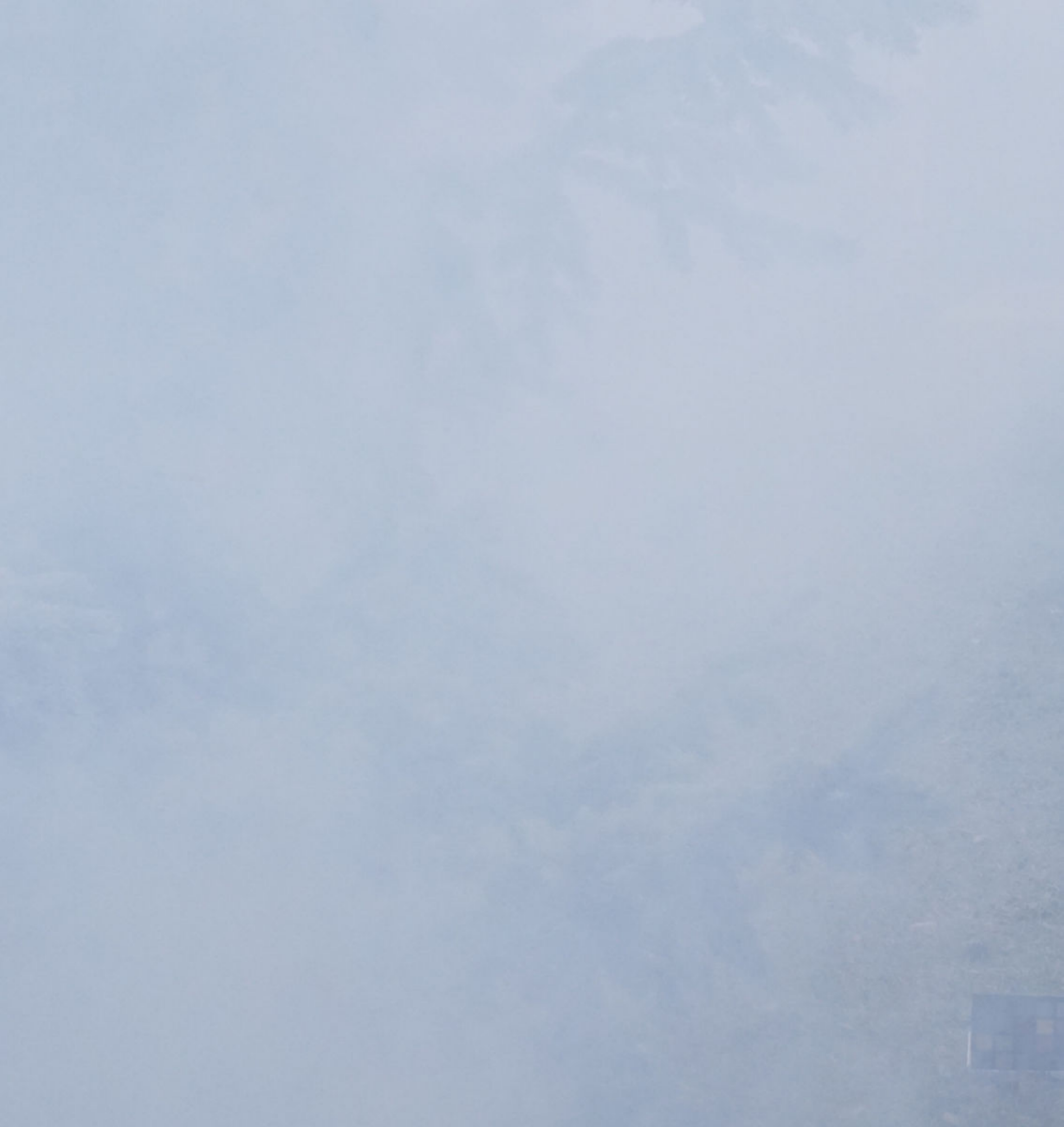}
  }
  \subfloat[MSBDN
  ]{
    \hspace{-0.1in}
    \includegraphics[width=0.293\columnwidth]{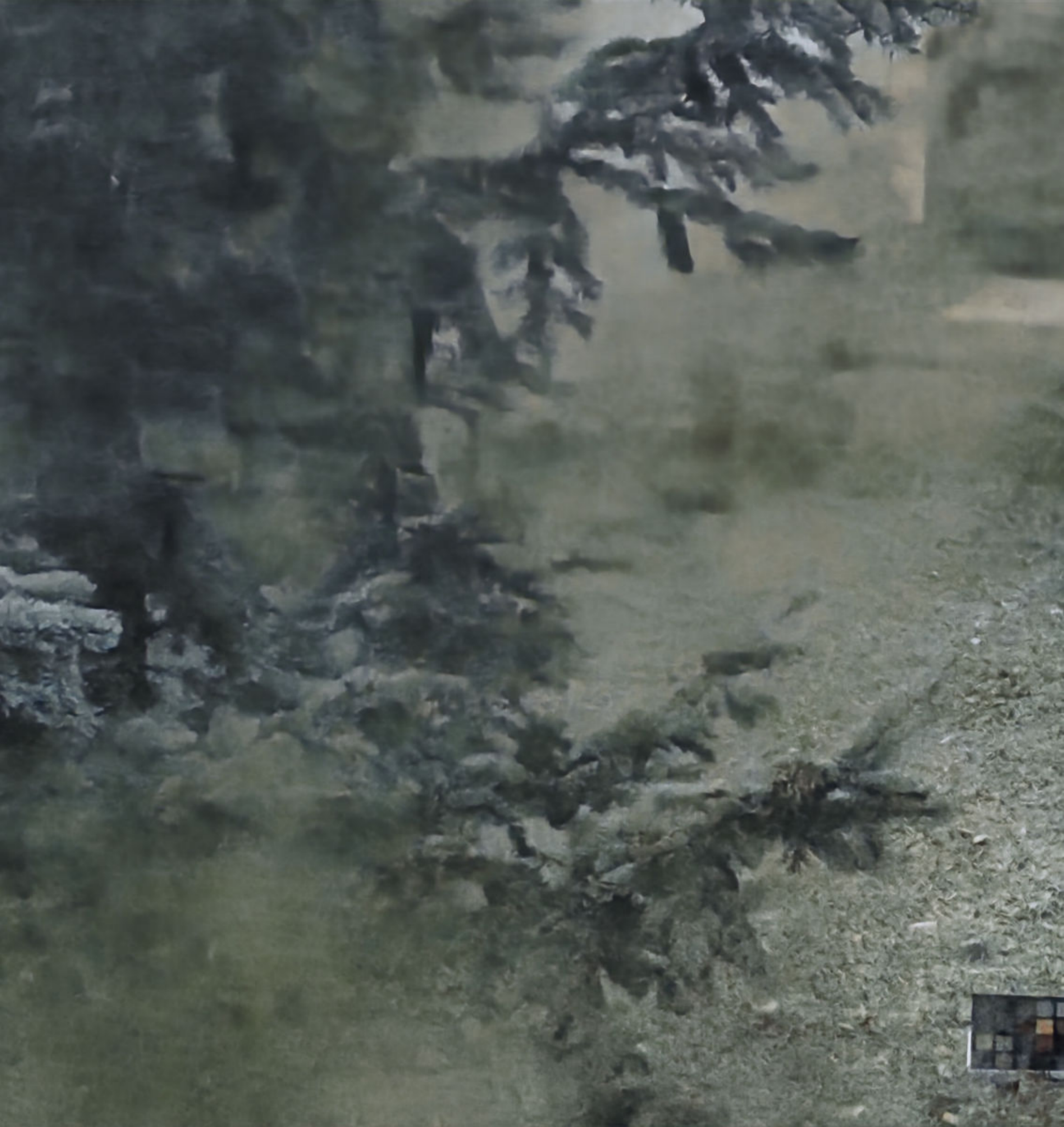}
  }
  \subfloat[FFA-Net 
  ]{
    \hspace{-0.1in}
    \includegraphics[width=0.293\columnwidth]{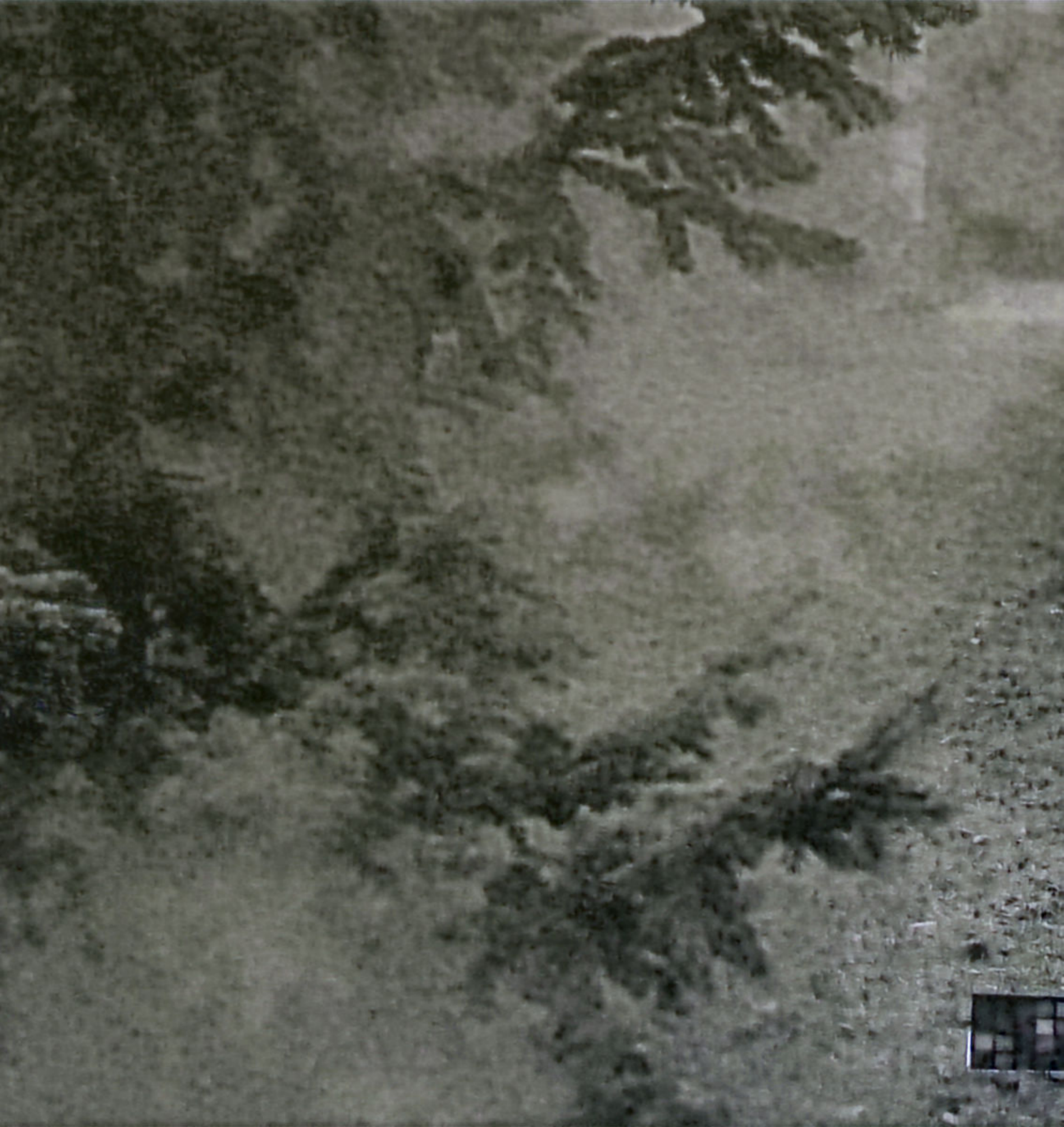}
  }
    \subfloat[SGID-PFF
  ]{
    \hspace{-0.1in}
    \includegraphics[width=0.293\columnwidth]{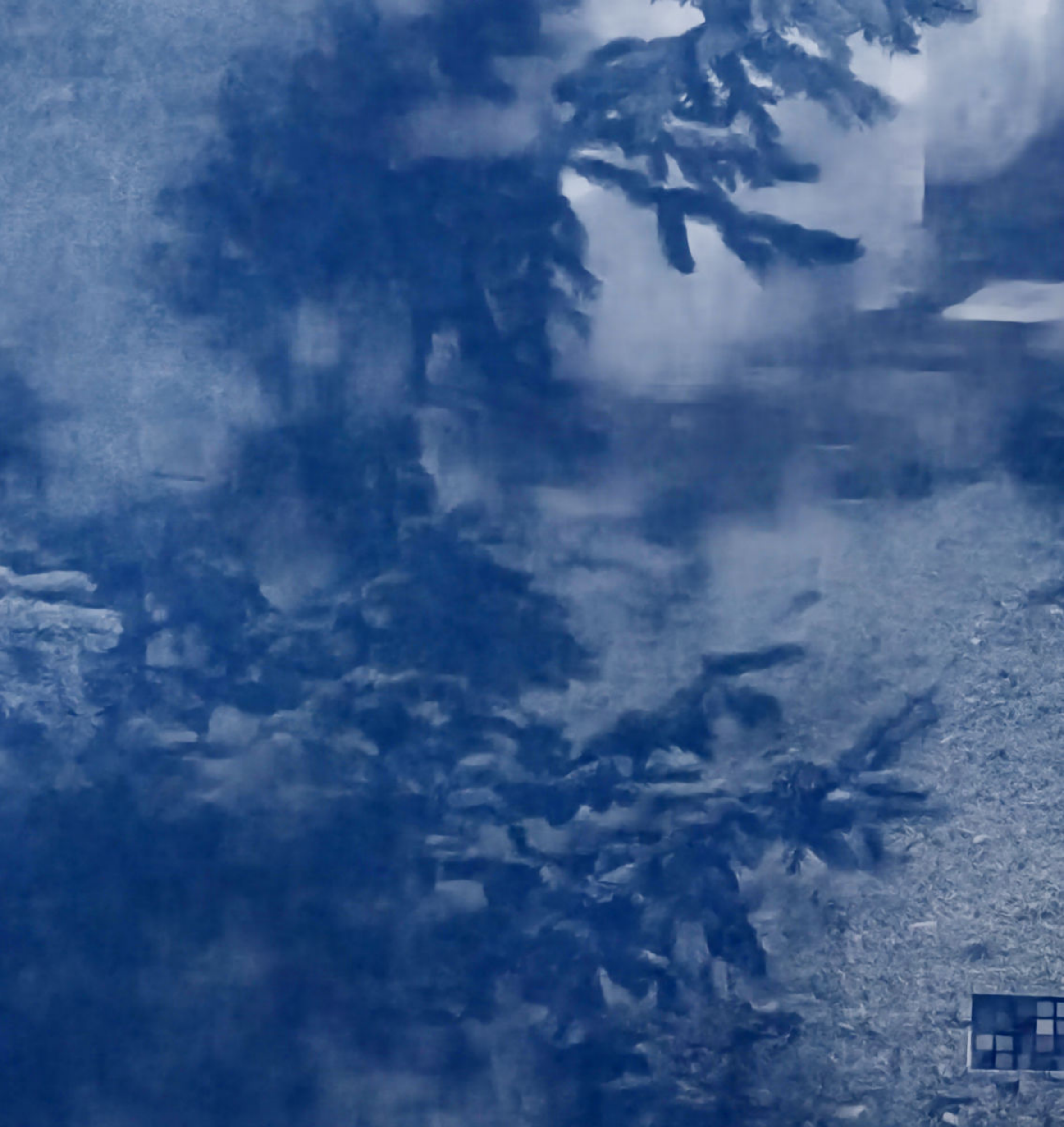}
  }
  \subfloat[Dehamer
  ]{
   \hspace{-0.1in}
    \includegraphics[width=0.293\columnwidth]{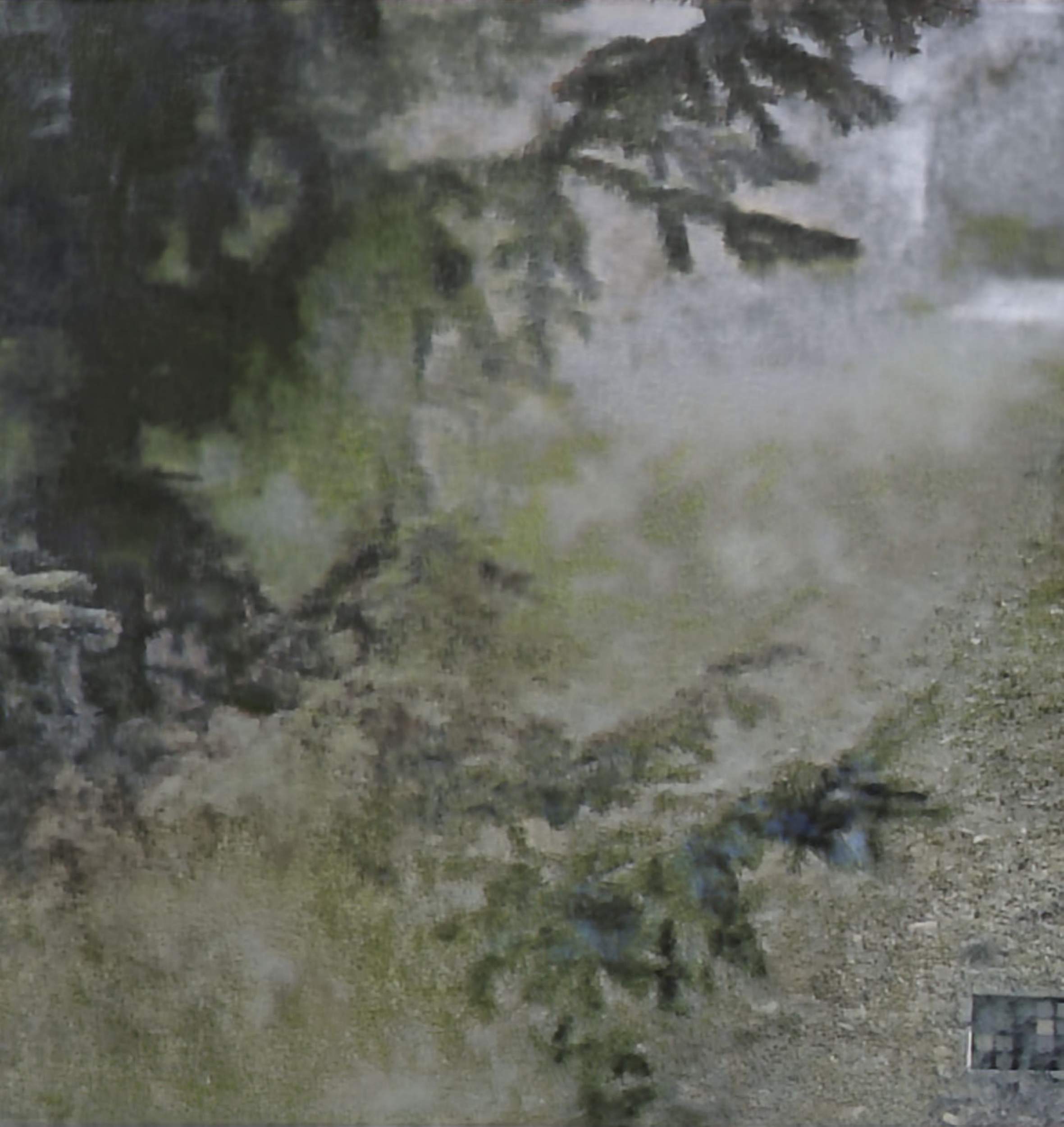}
  }
  \subfloat[Ours
  ]{
    \hspace{-0.1in}
    \includegraphics[width=0.293\columnwidth]{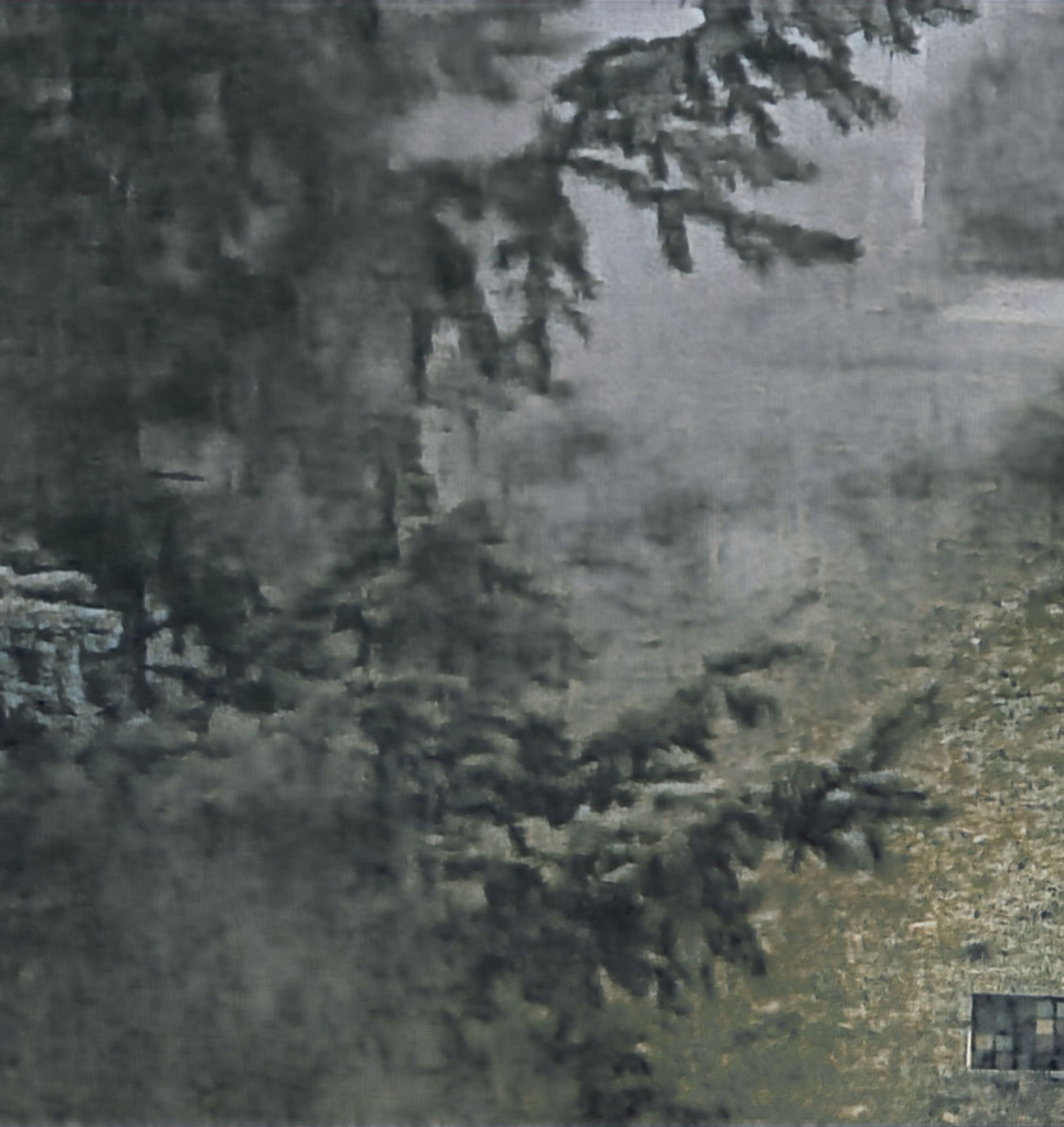}
  }
  \subfloat[GT
  ]{
    \hspace{-0.1in}
    \includegraphics[width=0.293\columnwidth]{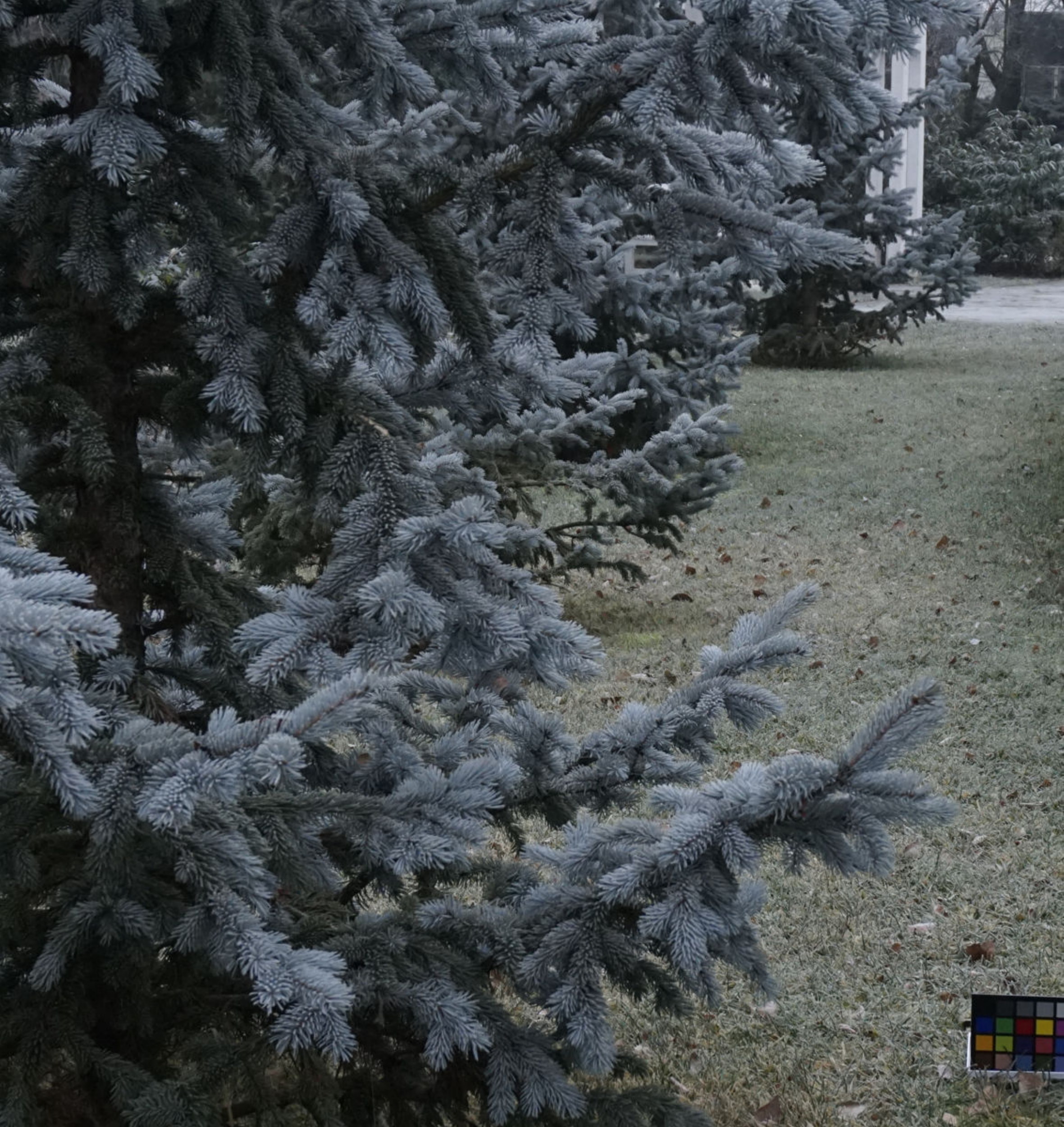}
  }
  \caption{\textbf{Visual comparisons on real hazy images.} The first row of images is from the O-HAZE, 
The second row of images is from the Dense-Haze. Our MB-TaylorFormer generates haze-free images with color fidelity and without artifacts. }
\vspace{-0.1cm}
    \label{fig:comparision_dense}
\end{figure*}

\subsection{Experiments on Synthetic Hazy Images}

     
     Table \ref{tab:compare} compares the performance of MB-TaylorFormer with the SOTA methods on synthetic datasets.
     Our basic model (MB-TaylorFormer-B) achieves $40.71$dB PSNR and $0.994$ SSIM on SOTS-Indoor. It improves the PSNR by $2.19$dB over the previous SOTA method SGID-PFF~\cite{bai2022self}, but with merely 
      $10\%$ number of parameters and $14\%$ computational cost of SGID-PFF. Furthermore, our large model (MB-TaylorFormer-L) achieves $4.12$dB gain over the SGID-PFF with approximately half of its complexity.
      Our method also surpasses the second best performing method Dehamer~\cite{guo2022image} by 2.91dB on the SOTS-Outdoor. 
      We also compare the visual results of MB-TaylorFormer-L with other SOTA dehazing methods. Fig.~\ref{fig:comparision} shows that the images generated by other methods are less natural in the shadows and high frequency regions. However, our MB-TaylorFormer generates haze-free images with better-restored details in the shadows, and is most similar to ground truth, especially in the high-frequency region.

\subsection{Experiments on Real Hazy Images}
    We further evaluate our MB-TaylorFormer against the SOTA methods on the real datasets O-HAZE~\cite{ancuti2018haze} and Dense-Haze~\cite{ancuti2019dense}. The quantitative comparison is shown in Table~\ref{tab:compare}. We have the following findings: 1) MB-TaylorFormer outperforms previous methods by up to $0.20$dB and $0.04$dB PSNR magnitude on the O-HAZE and Dense-Haze, respectively. 2) Usually, when training on small-size datasets, the non-convex losses of visual Transformer are expected to lead to poor performance~\cite{park2022vision}. However, MB-TaylorFormer still achieves the best PSNR and SSIM scores than other CNN-based models on small-size datasets.
    We also compare our MB-TaylorFormer with previous methods in terms of the visual quality of dehazed images, which are presented in Fig.~\ref{fig:comparision_dense}. 
    The dehazed images by MSBDN~\cite{dong2020multi} hardly remove the haze, FFA-Net~\cite{qin2020ffa} generates images with graininess and loss of detail, SGID-PFF~\cite{bai2022self} generates images with color distortion, and the result images by Dehamer~\cite{guo2022image} suffer from artifacts and texture loss. The haze-free images generated by our method are much cleaner.
\begin{table}[t]
\centering
\caption{\textbf{Ablation studies for the multi-scale patch embedding and multi-branch structure.} ``-S" means two convolutional layers in series with the same kernel size of 3, and ``-P" means two convolutional layers in parallel with the same kernel size of 3. Supplementary materials provide more details.}
\label{tab:tab2}
\scalebox{0.8}{
\begin{tabular}{c|c|cc|c|c}
\toprule
Branch & Type of Conv. & PSNR                            & SSIM                            & \#Params & MACs   \\ \hline
Single & Conv         & 38.27                           & 0.991                           & 2.655M   & 33.63G \\ \hline
\multirow{4}{*}{Double} & Conv-P         & 38.42                           & 0.991                           & 2.652M   & 37.89G \\
& Dilated Conv-P & 38.77                           & 0.991                           & 2.652M   & 37.89G \\
      & Conv-S         & 39.04                           & 0.992                           & 2.652M   & 37.89G \\
       & DSDCN-S         & \textbf{40.71} & \textbf{0.992} & 2.677M   & 38.51G  \\ \bottomrule
\end{tabular}}
\end{table}
\begin{table}[t]
\caption{\textbf{Effectiveness of multi-scale attention refinement module.} It balances computational burden and performance.}
\label{tab:tab4}
\scalebox{0.76}{
\begin{tabular}{c|cc|c|c}
\toprule
Methods                      & PSNR  & SSIM  & \#Params & MACs    \\ \hline
MB-TaylorFormer-B & 40.71 & 0.992 & 2.677M & 38.51G  \\
W/o MSAR                     & 38.74 & 0.990 & 2.582M & 37.60G  \\
$G: \mathbb{R}^{h\times 1\times H\times W}\to \mathbb{R}^{h\times C\times H\times W} $                      &          \textbf{40.88}  &   \textbf{0.993}    & 4.288M & 68.43G \\ \bottomrule
\end{tabular}}
\vspace{-0.2cm}
\end{table}

\subsection{Ablation Studies}

In this section, all MB-TaylorFormer models are trained on $256 \times 256$ pixel patches cropped from SOTS-Indoor. The epoch is set as $500$. Based on MB-TaylorFormer-B, we analyze the effectiveness of different modules on our framework. MB-TaylorFormer-L is used to explore the effects of branch dimension, channel dimension, and depth.

\begin{figure*}[t]
  \centering
  \captionsetup[subfloat]{labelformat=empty}
    \subfloat[Input
  ]{

    \includegraphics[width=0.6\columnwidth]{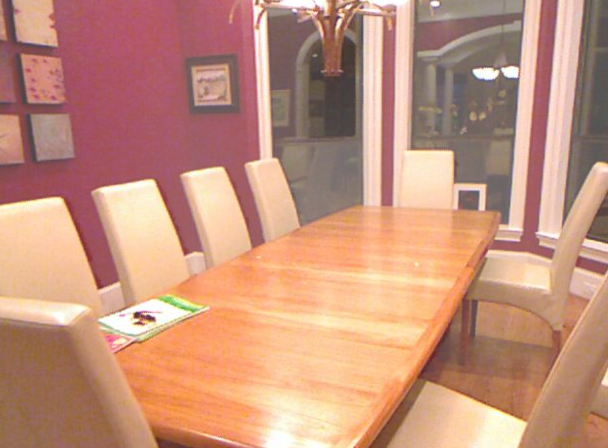}
  }
  \subfloat[Conv (Single)
  ]{
    \hspace{0.2in}
    \includegraphics[width=0.6\columnwidth]{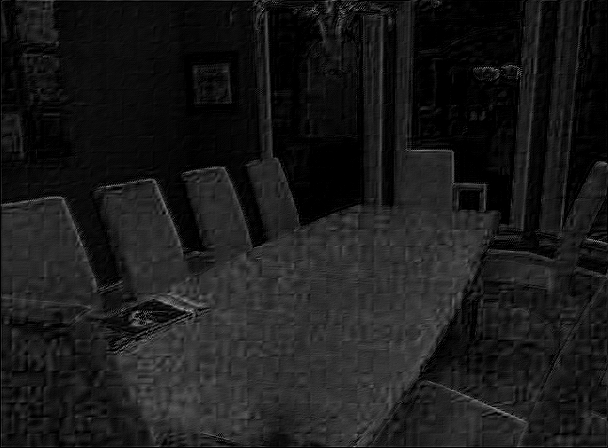}
  }
  \subfloat[DSDCN-S (Double)
  ]{
    \hspace{0.2in}
    \includegraphics[width=0.6\columnwidth]{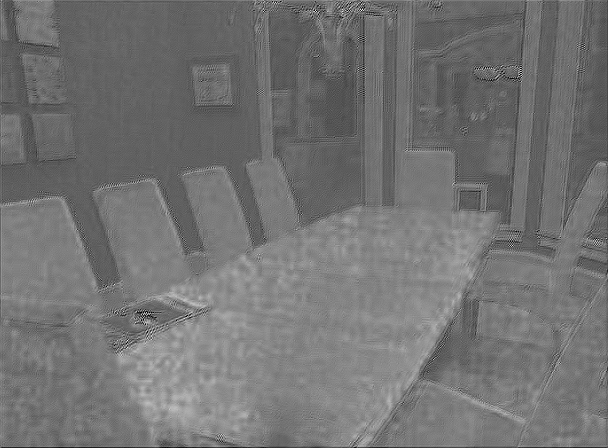}
  }
  \caption{\textbf{Comparison of feature map visualization for single-scale and multi-scale tokens.} The design of multi-scale and multi-branch can capture more powerful features.}
    \label{fig:aba_fig}
\vspace{-0.3cm}
\end{figure*}

\textbf{Study of multi-scale patch embedding and multi-branch structure.} In Table~\ref{tab:tab2}, we study the difference in the patch embedding and the different number of branches. Specifically, we set a single-branch model based on single-scale standard convolution as the baseline and modify it from the following aspects.
1) To examine the effect of multi-branch structure, we design the patch embedding on models with single-scale and multi-branching (Conv-P).
2) To study the effect of multiple receptive field sizes, we use parallel dilated convolutional layers (DF=1, 2) to embed patches (Dilated Conv-P). 3) To 
investigate the influence of multi-level semantic information, we further replace dilated convolution with standard convolution to embed patches, and employ the approach of connecting two convolutional layers in series (Conv-S). 4) To examine the effect of flexible receptive field shapes, we additionally replace standard convolution with DSDCN (DSDCN-S), as shown in Fig.~\ref{fig:pipeline}(b). 
The experiment shows that the performance from best to worst is DSDCN-S, Conv-S, Dilated Conv-S, Conv-P and Conv. This indicates that our multi-scale patch embedding can embed patches flexibly. As shown in Fig.~\ref{fig:aba_fig}, we visualize the average of the feature maps in the first stage of our network, and find that the multi-scale tokens (DSDCN-S) help the network to obtain richer information than the single-scale tokens (Conv). Thanks to the depth-separable design, our DSDCN has only a tiny increase in the number of parameters and computational cost compared to the depth-separable convolution.

\begin{table}[t]
 \centering
\caption{\textbf{Comparison with other linear self-attention modules.} We replace the T-MSA of our proposed model with another linear self-attention module.}
\label{tab:tab5}
\scalebox{0.76}{
\begin{tabular}{c|cc|c|c}
\toprule
Methods                     & PSNR  & SSIM  & \#Params & MACs   \\ \hline
MB-TaylorFormer-B & \textbf{40.71} & \textbf{0.992} & 2.677M & 38.51G \\
W/o MSAR                    & 38.74 & 0.990 & 2.582M & 37.60G \\\hline
T-MSA $\to$ MDTA~\cite{zamir2022restormer}                  & 38.57 & 0.991 & 2.582M & 35.20G \\
T-MSA $\to$ Swin~\cite{liang2021swinir}                  & 36.59 & 0.988 & 2.517M & 36.38G \\ 
T-MSA $\to$ TAR~\cite{katharopoulos2020transformers}                 & 36.74 & 0.987 & 2.517M & 34.00G \\ 
T-MSA $\to$ PVTv2~\cite{wang2022pvt}                 & 38.10 & 0.990 & 10.89M & 38.60G \\ 
T-MSA $\to$ Cswin~\cite{dong2022cswin}                 & 38.19 & 0.987 & 3.30M & 40.10G \\ 
T-MSA $\to$ LinFormer~\cite{wang2020linformer}                 & 36.12 & 0.983 & 48.70M & 371.68G \\ 
\bottomrule
\end{tabular}}
\end{table}

\begin{table}[t]
 \centering
\caption{\textbf{Analysis of approximation errors.} The smaller approximation error for softmax-attention, the better the performance.}
\label{tab:tab7}
\scalebox{0.8}{
\begin{tabular}{c|cc|c|c}
\toprule
Methods                      & PSNR  & SSIM  & \#Params & MACs    \\ \hline
Swin & \textbf{36.59} & \textbf{0.988} & 2.517M & 36.38G  \\
Swin + T-MSA-2nd                     & 36.50 & \textbf{0.988} & 2.517M & 36.38G  \\
Swin + T-MSA-1st                     &          36.37  &   0.987    & 2.517M & 36.38G \\ \bottomrule
\end{tabular}}
\vspace{-0.3cm}
\end{table}
\textbf{Effectiveness of multi-scale attention refinement module}. Table~\ref{tab:tab4} demonstrates our MSAR module provides a favorable gain of $1.97$dB over the counterpart without MSAR module, with only a tiny increase in the number of parameters ($0.1$M) and MACs ($0.905$G).
When we set the gating tensor $G \in \mathbb{R}^{h\times 1\times H\times W}  \to G \in \mathbb{R}^{h\times C\times H\times W} $, it provides only a tiny improvement in PSNR and SSIM, at the cost of a significant increase in the number of parameters and computation complexity. 
Taking into account both performance and computational burden, we set  $G \in \mathbb{R}^{h\times 1\times H\times W}$.

\textbf{Comparison with other linear self-attention modules}. Table~\ref{tab:tab5} compares the proposed T-MSA with several common linear self-attentive modules. 
Results show that the TaylorFormer has significant advantages over existing linear self-attention modules, which is attributed to the ability of T-MSA to model long-distance pixels and the approximation to the softmax-attention.

\textbf{Analysis of approximation errors.} To explore the approximation error and its impact, we investigate the effect of different orders of Taylor expansion for softmax attention. Considering that the associative law is not applicable to the second-order T-MSA (T-MSA-2nd), which leads to a significant computational burden, we perform first-order and second-order Taylor expansions for Swin. Table~\ref{tab:tab7} shows that T-MSA can effectively approximate softmax-attention, and the T-MSA-2nd is already very close to softmax-attention in performance. However, the computational complexity of both T-MSA-2nd and softmax-attention increases quadratically with image resolution, so it is difficult to model long-range pixel relationships in practical applications. Thus, we finally choose T-MSA-1st for our proposed method.



\begin{table}[]
\centering
\caption{\textbf{Results on the CSD dataset ~\cite{chen2021all} for snow removal.} The best and second best results are highlighted in bold and underlined, respectively.}
    \label{tab:CSD}
\scalebox{0.76}{

\begin{tabular}{c|cc|cc}
 \toprule
\multirow{2}{*}{Methods} & \multicolumn{2}{c|}{CSD (2000)}                                          & \multicolumn{2}{c}{Overhead}                                \\ \cline{2-5} 
                         & PSNR$\uparrow$                               & SSIM$\uparrow$                               & \multicolumn{1}{c|}{\#Param$\downarrow$} & MACs$\downarrow$ \\ \hline
DesnowNet~\cite{liu2018desnownet}                & 20.13                              & 0.81                               & \multicolumn{1}{c|}{26.15M}               & 1.7KG            \\
CycleGAN ~\cite{engin2018cycle}                & 20.98                              & 0.80                               & \multicolumn{1}{c|}{7.84M}               & 42.4G           \\
All in One~\cite{li2020all}               & 26.31                              & 0.87                               & \multicolumn{1}{c|}{44.00M}                 & 12.3G           \\
HDCW-Net~\cite{chen2021all}                 & 29.06                              & 0.91                               & \multicolumn{1}{c|}{6.99M}               & 9.8G            \\
TKL~\cite{chen2022learning}                      & 33.89                              & 0.96                               & \multicolumn{1}{c|}{31.35M}              & 41.6G           \\
SMGARN~\cite{cheng2022snow}                   & 31.93                              & 0.95                               & \multicolumn{1}{c|}{6.86M}               & 450.3G                  \\ 
Uformer~\cite{wang2022uformer}                   & 33.80                             & 0.96                               & \multicolumn{1}{c|}{9.03M}               & 19.8G                  \\
Restormer~\cite{zamir2022restormer}                   & \underline{35.43}                              & \underline{0.97}                              & \multicolumn{1}{c|}{26.10M}               & 141.0G                  \\\hline
Ours (-B) & \textbf{37.10} & \textbf{0.98} & \multicolumn{1}{c|}{2.68M}               & 38.5G            \\ \bottomrule
\end{tabular}}
\vspace{-0.2cm}
\end{table}

\begin{table}[]
\caption{\textbf{Results on the RainCityscapes dataset~\cite{hu2019depth} for rain removal.} ``-" indicates that the result is not available. The best and second best results are highlighted in bold and underlined, respectively.}
    \label{tab:rain}
    \centering
\scalebox{0.61}{
\begin{tabular}{c|cccccc|c}
\toprule
     & \begin{tabular}[c]{@{}c@{}}RESCAN\\ ~\cite{li2018recurrent}\end{tabular} & \begin{tabular}[c]{@{}c@{}}DAFNet\\ ~\cite{hu2019depth}\end{tabular}      & \begin{tabular}[c]{@{}c@{}}PCNet\\ ~\cite{jiang2021rain}\end{tabular} & \begin{tabular}[c]{@{}c@{}}EPRRNet\\ ~\cite{zhang2022beyond}\end{tabular} & \begin{tabular}[c]{@{}c@{}}Uformer\\ ~\cite{wang2022uformer}\end{tabular} & \begin{tabular}[c]{@{}c@{}}Restormer\\ ~\cite{zamir2022restormer}\end{tabular} & Ours(-B)      \\ \hline
PSNR$\uparrow$ & 28.97   & 30.06    & 24.49 & 31.11 & 32.26  & \underline{34.37}   & \textbf{36.55} \\
SSIM$\uparrow$ & 0.885   & 0.953  & 0.946 & 0.974  & 0.981 & \underline{0.988}   & \textbf{0.990} \\ \hline
\#Params$\downarrow$ & 0.15M                                                                                                                                                                       & -                                                                                                                                                                      & 0.63M                                                                                                                                                                      & 137.10M                                                                                                                                                                 & 9.03M                                                                                                                                                                   & 26.10M                                                                                                                                                                       & 2.68M                               \\
MACs$\downarrow$     & 9.7G                                                                                                                                                                       & -                                                                                                                                                                      & 2.4G                                                                                                                                                                      & 182.2G                                                                                                                                                                  & 19.8G                                                                                                                                                                   & 141.0G                                                                                                                                                                       & 38.5G                               \\\bottomrule
\end{tabular}}
\vspace{-0.5cm}
\end{table}
\subsection{The Generalization Capabilities}
%
Though this paper focuses on image dehazing, we have also evaluated the generalization capabilities of our proposed model on other tasks, specifically snow and rain removal, using the CSD~\cite{chen2021all} and RainCityscapes~\cite{hu2019depth} datasets. The setting of the training and testing dataset for snow removal follows the existing methods~\cite{chen2021all,cheng2022snow}, and the dataset setting for rain removal follows existing methods~\cite{hu2019depth, zhang2022beyond}.
%
%
%
The results, presented in Table~\ref{tab:CSD} and Table~\ref{tab:rain}, demonstrate that our model, MB-TaylorFormer, performs well in these tasks, indicating that its capabilities are not limited to dehazing. The supplementary material includes visualization results for these tasks.
%

\section{Conclusion}
In this paper, we propose a multi-branch linearized Transformer network, called MB-TaylorFormer, which consists of multi-scale patch embedding and Taylor expansion of self-attention. Multi-scale patch embedding with flexible shape of receptive field, multi-scale size receptive field, and multi-level semantic information, can enable flexible embedding of diverse visual tokens. The Taylor expansion of softmax-attention using the matrix multiplicative associate law reduces the computational complexity. Further, we correct the output of self-attention by gating attention, which allows MB-TaylorFormer to perform both long-range attention and local corrections.
%
%
Experimental results on various datasets demonstrate the effectiveness, lightness and generalization of the proposed MB-TaylorFormer.

\section{Acknowledgements}

This work was supported by the National Natural Science Foundation of China under Grant No. 62071500. Supported by Sino-Germen Mobility Programme M-0421. Supported by Guangdong Basic and Applied Basic Research Foundation under Grant No. 2023A1515012839. Supported by Shenzhen Science and Technology Program No.
JSGG20220831093004008.

{\small
\bibliographystyle{ieee_fullname}
\bibliography{egbib}
}

\end{document}